\documentclass[runningheads]{llncs}

\usepackage{eccv}

\usepackage{eccvabbrv}

\usepackage{graphicx}
\usepackage{booktabs}

\usepackage[accsupp]{axessibility}  %

\usepackage[dvipsnames]{xcolor}

\newcommand{\red}[1]{{\color{black}#1}}

\newcommand{\maplink}{\url{https://earthgen.github.io/}}

\definecolor{purple}{RGB}{160, 32, 240}
\definecolor{washblue}{RGB}{186, 224, 228}
\definecolor{sky}{RGB}{128, 128, 128}
\definecolor{seagreen}{RGB}{60, 179, 113}
\definecolor{building}{RGB}{128, 0, 0}
\definecolor{road}{RGB}{128, 64, 128}
\definecolor{sidewalk}{RGB}{0, 0, 192}
\definecolor{fence}{RGB}{64, 64, 128}
\definecolor{vegetation}{RGB}{128, 128, 0}
\definecolor{car}{RGB}{64, 0, 128}
\definecolor{sign}{RGB}{192, 128, 128}
\definecolor{pedestrian}{RGB}{64, 64, 0}
\definecolor{cyclist}{RGB}{0, 128, 192}

\def\be {\begin{equation}}
\def\ee {\end{equation}}
\def\beas {\begin{eqnarray*}}
\def\eeas {\end{eqnarray*}}
\def\bea {\begin{eqnarray}}
\def\eea {\end{eqnarray}}
\def\bes {\begin{equation*}}
\def\ees {\end{equation*}}

\newcommand{\bx}{\mathbf{x}}

\newcommand{\bI}{\mathbf{I}}

\newcommand{\bv}{\mathbf{v}}

\newcommand{\bu}{\mathbf{u}}

\newcommand{\bz}{\mathbf{z}}

\newcommand{\cL}{{\cal L}}

\newcommand{\cN}{{\cal N}}
\newcommand{\cU}{{\cal U}}

\usepackage{pgf}  %

\usepackage[breaklinks,colorlinks,citecolor=eccvblue]{hyperref}
\usepackage{epigraph}

\usepackage{orcidlink}

\begin{document}

\title{Generating the World from Top-Down Views} 

\titlerunning{Generating the World from Top-Down Views}

\author{Ansh Sharma\textsuperscript{*}%
\and Albert Xiao\textsuperscript{*}
\and Praneet Rathi%
\and Rohit Kundu 
\and Albert Zhai%
\and Yuan Shen%
\and Shenlong Wang%
}

\authorrunning{Sharma et al.}

\institute{University of Illinois Urbana-Champaign}

\maketitle

\def\thefootnote{*}\footnotetext{Equal Contribution}

\includegraphics[width=0.93\textwidth]{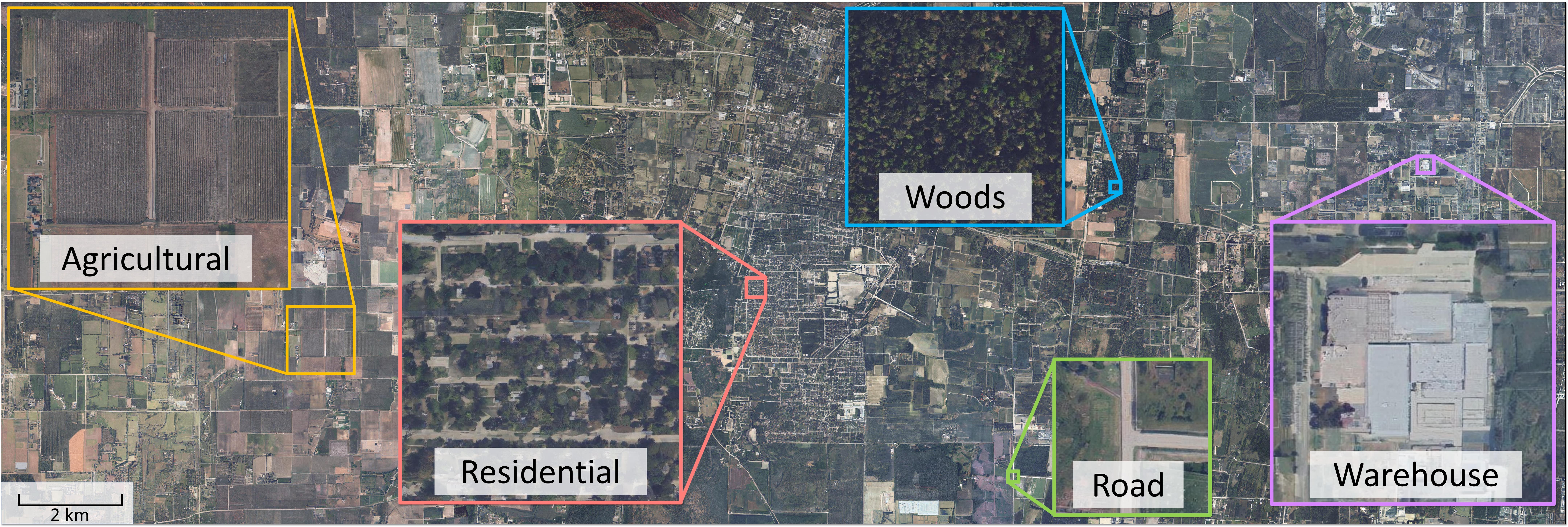}
            \vspace{-.5em}
    \captionof{figure}{EarthGen generates diverse and detailed terrains with clear and realistic features. The terrain spans 300 km$^2$ at 15cm/px, covering three Manhattans. The full 12 gigapixel output can be viewed interactively at \maplink}
    \label{fig:teaser}\vspace{1em}

\begin{abstract}
   In this work, we present a novel method for extensive multi-scale generative terrain modeling. At the core of our model is a cascade of superresolution diffusion models that can be combined to produce consistent images across multiple resolutions. Pairing this concept with a tiled generation method yields a scalable system that can generate thousands of square kilometers of realistic Earth surfaces at high resolution. We evaluate our method on a dataset collected from Bing Maps and show that it outperforms super-resolution baselines on the extreme super-resolution task of $1024 \times$ zoom. We also demonstrate its ability to create diverse and coherent scenes via an interactive gigapixel-scale generated map. Finally, we demonstrate how our system can be extended to enable novel content creation applications including controllable world generation and 3D scene generation.
  \keywords{Generative Modeling \and Remote Sensing} 
\end{abstract}

\section{Introduction}
\label{sec:intro}
\setlength{\parindent}{15pt}

There have been significant advancements in developing generative models capable of creating a range of visual data, including individual images~\cite{stable_diff, saharia2022photorealistic, ho2020denoising, song2019generative, karras2019style}, videos~\cite{singer2022make, guo2023animatediff, esser2023structure} and 3D objects~\cite{poole2022dreamfusion, li2023instant3d, chen2023fantasia3d}. These models have become pivotal in content creation, opening up numerous exciting applications.

However, the generation of large-scale representations of Earth's landscape is still a largely unexplored area. Successfully achieving this requires a balance of coherency, realism, and diversity. This is crucial because Earth's landscape is both highly structured and diverse on a macroscopic level, encompassing everything from vast plains to densely populated urban areas. At the same time, it is filled with rich, intricate details at the microscopic level. Any failure to capture both aspects can lead to inaccuracies and unrealism.

Current techniques for large-scale visual generation fall into two categories: the compositional approach and the hierarchical approach. The compositional approach~\cite{li2022infinitenature, shen2022sgam, xie2023citydreamer, lin2021infinity, chang2022maskgit, stable_diff} synthesizes unlimited images or 3D worlds by iteratively expanding the frontier, creating new local areas and seamlessly integrating them into the existing structure. This method excels in preserving high-resolution, fine-grained details and can generate images of any size. However, it struggles to accurately capture larger, macroscopic structures. In contrast, the hierarchical approach~\cite{saharia2022photorealistic, ho2022cascaded} uses a coarse-to-fine architecture. It starts by generating a basic layout and then progressively adds more detail in a manner akin to super-resolution. This method is effective in capturing high-level structures but is limited in the maximum size it can generate. This will be illustrated later through empirical results; for generating Earth observation data, existing methods in this category tend to lack sufficient detail at a finer level.

In this work, we present EarthGen, a novel framework for infinite-size, high-resolution earth observation imagery generation aimed at overcoming the aforementioned challenges. Our key insight is to combine the best of the worlds of hierarchical and compositional generation methods. Like hierarchical generation, we start the generative process at a significantly lower resolution than the final result. We then use a series of latent diffusion super-resolution modules to iteratively refine and introduce plausible features at each scale. Importantly, our super-resolution module is designed to be scale-aware. This ensures globally coherent outputs and captures realistic structures at multiple levels, ranging from regional to kilometer, meter, and centimeter scales. Simultaneously, diverging from conventional hierarchical methods, our method maintains high cohesiveness and expandability through a novel, diffuser-based composition method, allowing expansion to infinite sizes.

We demonstrate our framework's capabilities by generating gigapixel-scale terrain covering 30km x 10km, approximately three times larger than Manhattan, at a resolution of 15 cm/pixel as shown in Fig.~\ref{fig:teaser}. An interactive, full-resolution visualization of this terrain is available at \maplink. Our result showcases diversity, extreme realism at both macro and micro levels, and global coherency in structures and content. Additionally, we benchmark our approach against current state-of-the-art generation and super-resolution methods in the extreme generative task of $1024 \times$ zoom. Our experimental results demonstrate superior performance, both quantitatively and qualitatively, and are supported by a user study. We additionally showcase EarthGen's potential in controllable generation (conditioned on a map layout), as well as in 3D world creation (using an off-the-shelf depth reasoning module on top of EarthGen data).

Our contributions are listed as follows:
\begin{itemize}
    \item  We introduce a novel perpetual generation framework capable of creating realistic, arbitrarily sized visual images across up to 5 hierarchies, with a resolution difference of up to 1024 times.
    \item Building upon this generation method, we have developed EarthGen, a system capable of generating high-quality, large-scale earth observation images.
\end{itemize}

EarthGen builds upon advances in generative modeling techniques and systematically integrates them into the unique domain of earth observation data. This offers new opportunities to the broader communities of computer vision, remote sensing, environmental science, agriculture, and urban planning, paving the way for numerous applications ranging from asset creation for games to data augmentation and enhancement for earth observation, such as land cover classification and automatic mapping. We aim to assist researchers in addressing some of the most pressing environmental and societal challenges in the future.

\section{Related Works}
\label{sec:related}
\setlength{\parindent}{15pt}
\paragraph{Image-level Super-Resolution.}
Super-resolution models reverse the degradation process by recovering the high-frequency details from the low-resolution image~\cite{chen2022real}. Examples of degradation include noise injection and blurring. One idea for super-resolution is to explicitly approximate parameters for the degradation process~\cite{6751227, huang2020unfolding, gu2019blind} using prior assumptions for the degradation model. The challenge is that real-world degradation is too complicated to capture exactly. Alternatively, with synthetic and paired real-world datasets available, supervised~\cite{kohler2019toward, zhang2018residual, zhang2018residual} and self-supervised~\cite{shocher2018zero, 9423211, soh2020meta} approaches have been proposed to directly learn the mapping from low resolution to high resolution in a data-driven manner. Beyond learning the mapping like EDT ~\cite{li2021efficient} and HAT ~\cite{chen2023activating}, generative modeling is designed to produce additional details from low-frequency signals. For example, many recent works develop super-resolution frameworks with generative adversarial networks~\cite{lin2021infinity, wang2018esrgan, jo2020investigating, shang2020perceptual}. In particular, GigaGAN ~\cite{kang2023scaling} shows superior performance in $\times$32 upscaling while supporting text-conditioning. Meanwhile, diffusion models also work well for detail generation as can be seen in the cascaded diffusion modeling~\cite{ho2022cascaded} in Imagen~
\cite{saharia2022photorealistic} and DeepFloyd IF ~\cite{deep_floyd}. In this work, we seek to efficiently upscale to 1024 times relative to the input resolution, which is far beyond what existing work supports.

\paragraph{Unbounded Content Generation.}

Unbounded content generation seeks to produce 2D or 3D content at infinite scale while ensuring the fidelity of the generation. Perpetual video generation is one application that produces realistic and 3D-consistent fly-through videos. Researchers have achieved success in natural scenes in terms of fidelity and realism~\cite{liu2021infinite, li2022infinitenature, chai2023persistent}. Beyond perpetual view generation, prior attempts are capable of producing large-scale 2D maps~\cite{zhu2023mapprior} and 3D scenes ~\cite{lin2023infinicity, xie2023citydreamer, shen2022sgam}, even in complex urban settings. In our work, we demonstrate the ability to  generate city scale high-resolution remote sensing imagery. 
Generative Powers of 10~\cite{wang2023generativepowers} addresses a similar task of unlimited zoom via generation of a nested multi-resolution image using a pre-trained diffusion model. Our work differs in a few major ways: notably its factorizable formulation allows for tiling to densely super-resolve the entire base image rather than just a single pyramid, and the individually fine tuned layers ensure scale consistency and realism.

\paragraph{Diffusion Models.}
Before the recent success of deep learning-based diffusion model~\cite{sohl2015deep, song2019generative, song2021scorebased}, researchers were long puzzled by Generative Adversarial Networks (GANs), with issues like unstable training ~\cite{kodali2017convergence} and mode collapse~\cite{zhang2018convergence, thanh2020catastrophic}. While early attempts on score matching~\cite{hyvarinen2005estimation} were unable to produce high-fidelity images, the seminal work by Song \etal~\cite{song2019generative} greatly increased the capacity by introducing multi-scale noise perturbation in the process of Langevin Dynamics. This is further improved upon with modified schedulers~\cite{nichol2021improved}, non-Markovian denoising processes~\cite{song2020denoising}, and guidance techniques~\cite{dhariwal2021diffusion, ho2022classifier,radford2021learning,nichol2021glide}. More recently, Latent Diffusion Models~\cite{Rombach_2022_CVPR} separate perceptual compression through encoder-decoder structures, from semantic compression through diffusion models in the latent space, and introduce latent-space conditioning through cross-attention. 

With diffusion models exhibiting strongly customizable conditioning capabilities~\cite{stablewebui}, researchers have attempted to apply diffusion models to more complex composition and tile generation tasks. One proposed technique models diffusion models as energy-based models to compose objects for higher complexity generations~\cite{liu2022compositional}. Other works attempt to jointly de-noise over multiple spatial locations given a diffusion prior. For instance, Multidiffusion~\cite{bar2023multidiffusion} reconciles multi-tile denoising at each noise level by solving a least square problem, and Mixture of Diffusers~\cite{jimenez2023mixture} follows a similar idea but instead composes the noise predictions for each tile into a single prediction for the whole image.

\paragraph{Visual Foundation Models for Remote Sensing.}

Remote sensing images have received significant attention as a broad application within computer vision. Many previous remote sensing works have focused on classification~\cite{basu2015deepsat, lu2017remote, cheng2015effective, tuia2009active}, recognition~\cite{inglada2007automatic, penatti2015deep}, and segmentation~\cite{yuan2013remote, fan2009single}, as well as remote sensing-specific captions~\cite{lu2017exploring, qu2016deep, liu2022remote, luo2020improving, zhang2019description}. Recent works like GEO-Bench~\cite{earthbench} and Prithvi~\cite{prithvi} have advanced remote sensing data analysis through benchmarking and pre-trained foundation models, but have primarily focused on facilitating analysis rather than high-fidelity generative modeling. In contrast, we develop powerful generative models for synthesizing realistic remote-sensing imagery to unlock diverse use cases.

Prior work has explored image-to-image translation from map views to satellite images~\cite{shah2021satgan,espinosa2023generate}, with recent methods~\cite{espinosa2023generate} generating satellite views conditioned on input maps over regions like Scotland. However, leveraging detailed map priors eases the generation process by providing rich structural information. Our unconditional and super-resolution models learn to generate coherent global and local structures from much weaker or no conditioning, though we also demonstrate low-resolution map conditioning to loosely control terrain features.

DiffusionSat~\cite{diffusionsat} has also introduced utilizing diffusion models tailored for remote sensing data to support tasks like temporal generation, super-resolution, and inpainting. However, DiffusionSat as well as prior works for image-to-image remote sensing tasks primarily operate on generating single image tiles with a strong focus on conditioning methods for applications. In contrast, our tiled cascaded approach enables coherent large-scale remote sensing image synthesis without relying on dense conditioning for each tile.

Early techniques ~\cite{yang2015remote} for remote sensing super-resolution centered around learning mappings from low to high-resolution images, including sparse dictionaries ~\cite{zhang2013remote} and support vector regression ~\cite{zhang2011scale}. More powerful deep networks were recently compared on two constructed remote sensing datasets at $\times$2, $\times$4, and $\times$8 scales ~\cite{wang2022comprehensive}. Best performing models included CNN-based ~\cite{zhang2018residual}, attention-based ~\cite{haut2019remote, niu2020single}, and projection-based ~\cite{haris2020deep} methods. However, the stability and performance of these networks in extreme super-resolution contexts, requiring realistic feature hallucination, haven't been tested.

\section{Approach}
\setlength{\parindent}{15pt}
\begin{figure}[!t]
    \centering
    \includegraphics[width=\textwidth]{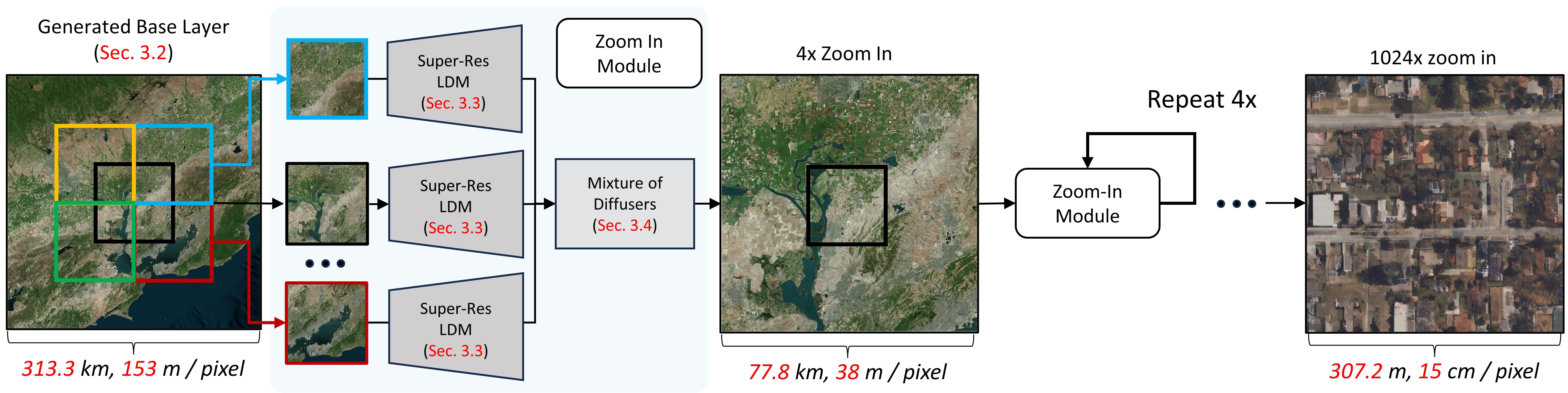}
    \caption{Our pipeline consists of a base layer module, cascaded super-resolution modules, and Mixture of Diffusers tiling.} %
    \label{fig:pipeline}
    \vspace{-5mm}
\end{figure}
We introduce EarthGen, a novel generative framework capable of continuously producing arbitrarily large-scale, photorealistic, coherent, and diverse landscapes from a bird's eye view, at resolutions as fine as 15cm per pixel. We approach this task as a cascaded and tiled generation problem (Sec.~\ref{sec:formulation}), comprising three key components: a latent diffusion-based generation capturing diverse structures at a regional scale (Sec.~\ref{sec:prior}); a coarse-to-fine multi-scale generative zooming module that progressively adds scale-aware visual details aided by negative conditioning (Sec.~\ref{sec:superresolution}); and a mixture model based sampling scheme ensuring cohesive outputs across spatially adjacent areas without detail loss (Sec.~\ref{sec:mixture}). We train our approach using a large-scale, multi-resolution satellite imagery dataset (Sec.~\ref{sec:training}).

\subsection{Problem Formulation}
\label{sec:formulation}

Formally, let $\bx: \cU \times \mathcal{S} \rightarrow \mathbb{R}^{3}$ represent a multi-resolution spatial RGB map defined in a geo-coordinate system. We denote $\bx_{\mathbf{u}, s} \subset \mathbb{R}^3$ as the sampled RGB appearance at region $\mathbf{u} \in \cU \subset \mathbb{R}^2$ at scale $s \in \mathcal{S}$. Our goal is to model this functional's distribution with a generative probabilistic model $p_\theta(\mathbf{x})$, where $\theta$ represents learnable parameters. This would enable us to sample multiple, diverse virtual worlds from top views, representing different landscapes, appearances, and zoning. Modeling this distribution poses two main challenges: 1) The mapping can extend indefinitely in the 2D geo-coordinate space, creating an infinite data generation problem; 2) Earth observation imagery captures a wide range of detailed structures at different scales, adding to the complexity. These challenges lead us to factorize the joint distribution into a series of conditional distributions:
\begin{align} \label{eq:1}
  p_{\theta}(\mathbf{x}) &= \prod_{\mathbf{u} \in \mathcal{U}} \underbrace{p_{\theta}(\mathbf{x}_{\mathbf{u}, 0})}_\mathrm{base} \prod_{s \in \mathcal{S} / 0 } %
  \underbrace{p_{\theta} (\mathbf{x}_{\bu, s} | \mathbf{x}_{s-1})}_\mathrm{zoom-in} %
\end{align}
where $p_{\theta}(\mathbf{x}_{\bu, 0})$ is the prior unconditional distribution of the base model, $p_{\theta} (\mathbf{x}_{\bu, s} | \mathbf{x}_{s-1})$ is a zooming module that models the conditional distribution of the visual appearance at scale $s$ given its preceding layers data $\bx_{s-1}$ at scale $s-1$. Factorization across scales and spaces allows us to {\it sample efficiently} and {\it capture structured details across each scale}. 

\red{To make coherent large-scale generation feasible, we further decompose each layer into a grid of overlapping tiles.} Inspired by ~\cite{hintonPOE, Du2020COM, liu2022compositional}, we then formulate this distribution as a compositional product of experts: %
\begin{equation}
\label{eq:2}
    p_{\theta} (\mathbf{x}_{\bu, s} | \mathbf{x}_{s-1}) \propto \overbrace{\prod_{\bv \in \cN_{\bu, s-1}} {\underbrace{p_{\theta} (\mathbf{x}_{\bu, s} | \mathbf{x}_{\bv, s-1}, \neg \mathbf{e}^-)}_\textrm{superres}}^{w_{\bu, \bv}}}^\mathrm{tiling}
\end{equation}
\red{where $\bu$ is the region-of-interest at scale $s$, $\cN_{\bu, s-1}$ is the set of tiles at layer $s-1$} that overlap with $\bu$, 
$\mathbf{e}^-$ is a negative text prompt embedding,
and $w_{\bu, \bv}$ is \red{a normalized gaussian mixing weight}. 
The core idea of this design is to ensure {\it coherency across scales} and {\it coherency between spatially neighboring areas}. 

Fig.~\ref{fig:pipeline} depicts the overall framework of our approach with a highlight on each module. We now describe each module's parameterization and how we sample from it.

\subsection{Unconditional Base Layer Model}
\label{sec:prior}

We begin our system with a base layer generation at low resolution. In particular, we sample $\bx_0 \sim p_\theta(\bx_0) = \prod_{\bu \in \cU} p_\theta(\bx_{\bu, 0})$ by modeling the data distribution via a latent diffusion model ~\cite{Rombach_2022_CVPR}. Latent diffusion models learn to create data by reversing a diffusion process within a latent space induced by a variational autoencoder.

Within the latent space, the LDM's forward diffusion process is modeled as a Markov chain that corrupts latent $\mathbf{z}_0^0 = \text{enc}(\mathbf{x}_0)$ into noise over \(T\) steps. At each step \(t\), noise is added according to $
\mathbf{z}_0^t = \sqrt{\alpha_t} \mathbf{z}_0^{t-1} + \sqrt{1 - \alpha_t} \mathbf{\epsilon}$, where $\mathbf{\epsilon} \sim \mathcal{N}(\mathbf{0}, \mathbf{I})$ and $\alpha_t$ is a variance schedule. 

The reverse process is learned. From a score model perspective, the LDM predicts the gradient of the data's negative log-likelihood: $
    \epsilon_\theta(\bz_0^t, t) \propto -\nabla \log p_\theta (\text{dec} (\bz_0^t)) $

We can sample the previous time step as follows:
\begin{equation} \label{eq:4}
    \bz_0^{t-1} = \frac{1}{\sqrt{\alpha_t}} \left( \bz_0^t - \frac{1 - \alpha_t}{\sqrt{1 - \overline{\alpha}_t}} \epsilon_\theta(\bz_0^t, t) \right) + \sigma_t \epsilon,
\end{equation} 
where $\sigma_t$ is a noise term to add stochasticity, $\epsilon \sim \cN(\mathbf{0}, \bI)$, and $\overline{\alpha}_t$ is the cumulative product of $\alpha_1$ through $\alpha_t$. Starting with $\bz_{T} \sim \cN(\mathbf{0}, \bI)$, we can generate our base layer by iteratively sampling until $\bz_0^0$, after which the decoder brings the latent back to the image space by $\mathbf{x}_{0} = \text{dec}(\mathbf{z}_0^0)$.

\subsection{Cascaded Generative Super-Resolution}
\label{sec:superresolution}

The core of our system centers around a set of cascaded super-resolution modules, each trained independently to specialize in hallucinating features at a given resolution. Each module learns $p_\theta(\bx_{\bu, s} | \bx_{\bu, s-1})$ via an LDM. By passing in a base image sequentially through the cascade, we iteratively upscale it to the desired resolution while introducing realistic hallucinated features at each stage. This formulation takes inspiration from Imagen ~\cite{saharia2022photorealistic}, which generates high-fidelity images by a low-resolution base model and two subsequent super-resolution models. 

\paragraph{Super Resolution Architecture.}  
We use the Stable Diffusion x4 Upscaler~\cite{Rombach_2022_CVPR} as the starting point for each super-resolution model, allowing us to increase the resolution by a factor of 4 during each step of the cascade. The model consists of a VAE encoder/decoder which maps RGB images of dimension $m \times n \times 3$ into a latent space with new dimension $\frac{m}{4} \times \frac{n}{4} \times 4$. This results in the latent space having the same spatial dimensions as the low-res image, allowing us to concatenate the two into a 7-channel input to the denoising U-nets at each time step. We can then learn super-resolution via an LDM conditioned on the low-resolution tile, where the model learns to estimate
$\mathbf{\epsilon}_\theta(\mathbf{z}^t_{\mathbf{u}, s}, t, \mathbf{x}_{\mathbf{u}, s-1}) \propto -\nabla \log p_\theta(\bz^t_{\bu, s} | x_{\bu, s-1})$.

\paragraph{Negative Conditioning.} 
 However, sampling directly using Eq.~\ref{eq:4} with this formulation leads to significant issues with compounding errors over layers. Ho \textit{et al.}~\cite{ho2022cascaded} addresses this issue by conditioning each stage on a single class label. \red{However, due to the scale of our task, direct conditioning alone would not be sufficient. Densely labeling each tile would be prohibitively expensive, while using a shared label would reduce the diversity of the model's generations across scales/locations.}

Instead, inspired by recent progress in compositional text conditioning ~\cite{stablewebui, liu2022compositional}, we utilize inference-time negative text conditioning to direct our images away from low-quality outputs, for instance, blurry or deformed images, based on the model's pre-trained priors. 
In particular, given a fixed negative prompt embedding $\mathbf{e}^-$, we seek to predict \begin{equation}\label{eq:5} p_\theta(\bx_s |\neg \mathbf{e}^{-}, \bx_{s-1}) \propto \frac{p_\theta(\bx_s|\bx_{s-1})^{\red{2}}}{p_\theta(\bx_s | \mathbf{e}^-, \bx_{s-1})}\end{equation}

Because the noise estimates model the negative log-likelihood of the data, by adding a hyperparameter to control the strength of the negative conditioning, we can define a corresponding modified noise prediction function to use in our super-resolution modules as:
\begin{equation}
\label{eq:6}
 \overline{\epsilon}_\theta(\mathbf{z}^t_{\mathbf{u}, s}) =
 \mathbf{\epsilon}_\theta(\mathbf{z}^t_{\mathbf{u}, s}) +  \lambda_\mathrm{neg}\left(\mathbf{\epsilon}_\theta(\mathbf{z}^t_{\mathbf{u}, s}) - \mathbf{\epsilon}_\theta(\mathbf{z}^t_{\mathbf{u}, s}, \mathbf{e}^-)\right)   
\end{equation}

Note that we left out the common conditioning parameters ($\bx_{\bu, s-1}, t$) for readability. \red{Derivation details are provided in Appendix \ref{sec:derivations}. This formulation allows us to avoid low-quality outputs from compounding errors without dense labels while maintaining global diversity.}

\subsection{Mixture of Diffusers for Tiling Consistency}
\label{sec:mixture}

Directly modeling the zoom for each tile independently in Eq. \ref{eq:1} would lead to tile border inconsistencies. Markov Random Fields (MRFs) or sequential modeling could be used to explicitly enforce conditioning between tiles at a given layer, but both are expensive to sample from. %

Alternatively, we integrate a mixture of diffusers inspired by ~\cite{jimenez2023mixture}, where each tile is conditioned on multiple overlapping views from the preceding scale as visualized in Fig. \ref{fig:pipeline}. This modifies the diffusion process to make the noise prediction at each time step a Gaussian blend of the predictions for that region from each tile which covers it in part. We now define a tiled noise as follows:
\begin{equation}
\hat{\mathbf{\epsilon}}_{\theta}(\bz_{\bu, s}^t, t, \bx_{\bu, s-1}) = \sum_{\bv \in \cN_{\bu, s-1}} w_{\bu, \bv} \cdot \mathcal{T}_{\bu, \bv} \left(\overline{\epsilon}_\theta(\bz_{\bv, s}^t, t, \bx_{\bv, s-1})\right),
\end{equation}
\red{where $\cN_{\bu, s-1}$ is the neighborhood of tiles at resolution $s-1$ which overlap with $\bu$, $\mathcal{T}_{\bu, \bv}$ is a translation operator which zero pads and shifts the predicted noise from view $\bv$ to $\bu,$ and $w_{\bu, \bv}$ is a normalized blending matrix proportional to a Gaussian centered at $\bv$}. Noting that the predicted noises learn to model the negative log likelihood of the data, this leads us directly to the tiling component mentioned in Eq. \ref{eq:2}.

To sample, we aggregate all view estimates into a single image-level estimate $\hat{\epsilon_\theta}(\bz^{t}_s, t, \bx_{s-1})$ where each pixel is the weighted average of the predictions for it from each view containing it. Then, we can sample the previous timestep for the entire image simultaneously using Eq. \ref{eq:4} with $\hat{\epsilon_\theta}$.

The denoised latent is decoded via tiled decoding~\cite{stablewebui}, which involves extracting overlapping patches from the latent, decoding them separately, and blending linearly to merge and get our final output for the layer $\bx_s$.

\subsection{Training}
\label{sec:training}

\paragraph{VAE Tuning.} We observed an inability of the pre-trained VAE to model satellite imagery well, especially at lower zooms. To alleviate this, we began by fine-tuning our VAE across all resolutions in the dataset via an image reconstruction task. We optimize a composite loss function comprising three terms: %
\begin{equation}
\cL_{VAE}(\phi; \bx) = \mathcal{L}_{\text{MSE}}(\phi; \mathbf{x}) + \lambda_{\mathrm{KL}}  \mathcal{L}_{\text{KL}}(\phi) + \lambda_{\mathrm{LP}} \mathcal{L}_{\text{LPIPS}}(\phi; \mathbf{x})
\end{equation}
where \( \mathbf{x} \) denotes the high-resolution target image, \( \phi \) represents the parameters of the VAE, 
\( \mathcal{L}_{\text{MSE}} \) is the mean-squared error, \( \mathcal{L}_{\text{KL}} \) is the KL divergence that regularizes the latent space, \( \mathcal{L}_{\text{LPIPS}} \) measures the perceptual similarity, 
and \( \lambda_{\mathrm{KL}}, \lambda_{\mathrm{LP}}\) are weight hyperparameters.

\paragraph{LDM Training.}

After freezing the VAE, we train an LDM over a single resolution of data to unconditionally model the map's base zoom layer. Training involves learning \(\theta\) by optimizing a variant of the evidence lower bound (ELBO). Ho \textit{et al.}~\cite{ho2020denoising} show that this is empirically equivalent to minimizing the mean squared error between the predicted noise and actual noise, allowing optimization via:
\begin{equation}
\label{eq:LDM}
\mathcal{L}(\theta) = \mathbb{E}_{\mathbf{z}_0, \mathbf{\epsilon}, t} \left[ \| \mathbf{\epsilon} - \epsilon_\theta(\mathbf{z}_t, t) \|_2^2 \right]
\end{equation}

\paragraph{Super-Resolution Training.} Using the same frozen VAE, each component of the cascade is trained independently on paired low/high-resolution data, each with a $\times$4 resolution gap. The training objective at scale $s$ only differs from Eq. \ref{eq:LDM} due to conditioning on the corresponding scale $s-1$ tile:
\begin{equation}   
\mathcal{L}_{\mathrm{sr}}(\theta, s) = \mathbb{E}_{\mathbf{x}_{\mathbf{u}, s}, \mathbf{\epsilon}, t} \left[ \| \mathbf{\epsilon} - \epsilon_\theta(\mathbf{z}^t_{\mathbf{u}, s}, t, \mathbf{x}_{\mathbf{u}, s-1}) \|_2^2 \right]
\end{equation}

\setlength{\parindent}{15pt}

\section{Experiments}

\paragraph{Dataset.} 
\label{sec:dataset}
We construct a custom dataset via the Bing Maps API~\cite{bingmaps} by querying satellite images at varying zooms and coordinates. The maximum zoom available is denoted as zoom 20, corresponding to 15cm/px, with each decrease in level corresponding to a $\times$2 reduction of resolution. We randomly sample concentric image pyramids of size $2048 \times 2048$ at zooms ranging from 10 to 20 over latitudes from -66 to 66 to minimize projection distortion.

We observe resolution availability falls into three main classes.
The coarsest class has resolutions up to Bing's zoom 13, corresponding to 19.2 m/px at the equator. Most of these occur over the oceans, so due to the limited resolution and lack of diversity, we ignore this class in constructing our training dataset. The remaining data over land tends to have resolution availability up to zoom 20 (15 cm/px) over the USA and Western Europe, and up to zoom 19 (30 cm/px) over the rest of the world. We compile 32,000 image stacks available up to zoom 19 (Bing19), and an additional 25,000 available up to zoom 20 (Bing20).

We additionally sample another dataset to prioritize urban locations which we call BingUrban. Using a list of the top 1000 most populated cities in the US~\cite{USCities} and the 36 most populated European cities~\cite{EUCities}, we collect 12,000 image stacks with zoom 20 availability centered at these cities with up to a 5-kilometer jitter in any direction.

We evaluate our models on two validation sets, each with roughly 100 $2048 \times 2048$ image pyramids, sampled similarly to Bing20 and BingUrban respectively.

\paragraph{Implementation Details.}

We use the Stable Diffusion $\times$4 Upscaler ~\cite{Rombach_2022_CVPR} architecture as our base for each super-resolution module. The model is trained to super-resolve inputs by a factor of 4 and takes in a \red{concatenation of the low-resolution image and the latent representation at each step.}  

We initially fine-tune a shared VAE across all resolutions with hyperparameters $\lambda_{\mathrm{KL}} = 10^{-9}, \lambda_{\mathrm{LPIPS}} = 0.1$. Freezing the VAE weights, we then tune each super-resolution module for 100k steps of batch size 24. Each $\times$4 model, referenced as 10to12, 12to14, etc. up to 18to20, is trained independently on all data stacks which contain its higher resolution. We train on high-resolution/low-resolution pairs with side lengths of 512px and 128px respectively.

At inference time, we utilize Mixture of Diffusers with a latent tile size of 128 and stride 64. We set $\lambda_\mathrm{neg}$ to 5, 2, 3, 3, and 4 for the models from 10to12 to 18to20 respectively on the prompt ``blurry, low res, low quality''. These parameters were decided experimentally to attain a good balance between consistency and realism over both urban and non-urban data. For VAE tiled decode, we use latent sliding window size 512 and tile overlap coefficient 0.25.

We utilize Adam~\cite{KingBa15} as our optimizer for all stages with learning rate 1e-06 and use DPMSingleStepSolver~\cite{lu2022dpm} with 50 time-steps as our scheduler during inference for the unconditional and super-resolution modules.

\paragraph{Metrics.}
Frechet Inception Distance (FID)~\cite{heusel2017gans} and Kernel Inception Distance (KID)~\cite{binkowski2018demystifying} are metrics for the quality of generated images with regard to a target distribution, which we use as measures of realism. Both are evaluated over 2048 features, and we use subset size 1000 with 100 subsets for KID.

Each validation set contains $4900$ $512 \times 512$ tiles, generated by super-resolving 100 zoom 10 ground truth images at $512 \times 512$ pixels to zoom 20 by alternatingly super-resolving and center cropping. The final super-resolved $2048 \times 2048$ output is split into 49 overlapping tiles to be used as inputs for the metrics in order to appropriately penalize any tiling artifacts. A separate identically sized dataset of ground truth images is used for calculating metrics.

\subsection{Image Super-Resolution}
\label{sec:results}

\paragraph{Baselines.}
We compare EarthGen to four state-of-the-art models, Real-ESRGAN~\cite{wang2022realesrgan}, HAT~\cite{chen2023activating}, LIIF~\cite{chen2021learning}, and the base Stable Diffusion Upscaler~\cite{Rombach_2022_CVPR}. Real-ESRGAN is fine-tuned on our dataset for more steps and equal time as our models for fair comparison. HAT and LIIF remain untuned because these models are regressive rather than generative, and as such would be unable to introduce new terrain and manmade features even with fine-tuning. We additionally evaluate bicubic interpolation as a baseline for contextualization. 

\paragraph{Quantitative Results.}

Table \ref{tab:example} provides quantitative results from our evaluated metrics. EarthGen exhibits significant improvements in both FID and KID and in both the general and urban validation set. This shows that the quality of the generated images is significantly improved compared to existing models.
\begin{table}[!htbp]
  \centering
  \begin{tabular}{ccccc}
    \toprule
    Method & FID($\downarrow$) & KID ($\downarrow$) \\
    \midrule
    SD x4 Upscaler~\cite{Rombach_2022_CVPR} & 163.01~/~223.95 & 0.0915~/~0.1594   \\
    Real ESRGAN~\cite{wang2022realesrgan} & 160.06~/~231.85 & 0.0858~/~0.1824   \\
    HAT~\cite{chen2023activating} & 364.40~/~455.02 & 0.3371~/~0.4821  \\
    LIIF~\cite{chen2021learning} & 396.32~/~483.56 & 0.3857~/~0.5249   \\
    Interpolation & 344.11~/~414.59 & 0.2965~/~0.4158   \\
    \midrule
    \textbf{Ours} & \textbf{66.40~/~91.04} & \textbf{0.0210~/~0.0533}\\
    \bottomrule
  \end{tabular}
  
  \vspace{5pt}
  
  \caption{General/Urban $\times$1024 Super-Resolution Metrics. Our method strongly outperforms all baselines in all metrics/datasets.\vspace{-15pt}}
  \label{tab:example}
\end{table}

\paragraph{Qualitative Results.}

Figure \ref{fig:qual_comp} provides a qualitative comparison of each model. Although every model produces comparable results in the first $\times$4 superresolution, it quickly becomes clear that our model produces the best images by $\times$1024 superresolution with respect to realism, quality, and consistency. 

\begin{figure}[!htbp]
  \centering
  \vspace{-5mm}
  \includegraphics[scale=0.23]{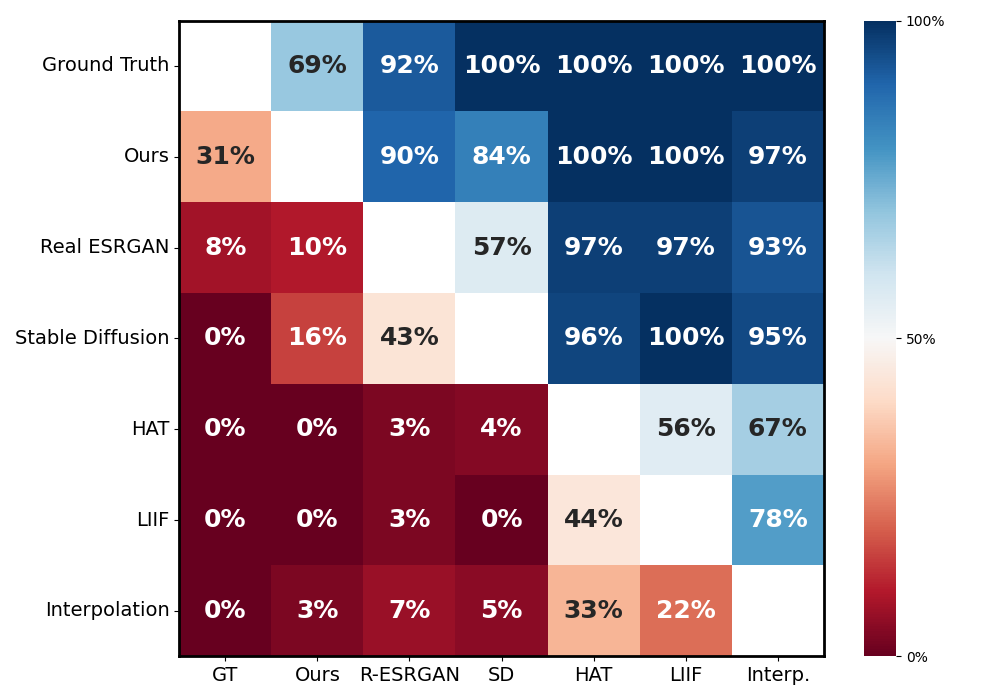}
\caption{\vspace{-1mm}Pairwise win-rates between models based on user study. Each cell represents the row's win rate against the column.\vspace{-2mm}} 
\label{tab:user-study}
\end{figure}

For Stable Diffusion, the compounding effects of low-quality generations have made the final image unrealistic and low quality despite consistency with previous layers. Real ESRGAN can generate high-quality and consistent images across all levels but fails to produce an urban layout with buildings and roads, despite the strong urban layout seen even at $\times$4 superresolution. All three of HAT, LIIF, and basic Bicubic Interpolation are incapable of hallucinating new features, and their final outputs merely look like a smoothened version of the 4 corresponding pixels in the original image. Finally, EarthGen succeeds in all three categories as it consistently preserves the urban layout, has vastly better quality than all except Real ESRGAN, and has easily discernable structures such as roads, houses, trees, and more.
\paragraph{User Study.}
We additionally conducted a user study to compare each baseline method with ours and the ground truth. We presented users with \red{unlabeled} one-on-one match-ups between two 2048$\times$2048 zoom 20 images generated by super-resolving zoom 10 images with different methods \red{alongside the following instructions: \textit{You will be
presented with a number of pairs of aerial photos. Please
select the one which looks better in terms of realism and coherency}}. 30 \red{volunteer student} participants engaged in the study aggregating 522 total votes. We observe in Fig. \ref{tab:user-study} that our method outperforms all but the ground truth, against which it still wins around 30\% of the time.

\begin{table*}[]
    \label{fig:baselines} 
    \centering
    \resizebox{\linewidth}{!}{
\setlength{\tabcolsep}{0.2em} %
\renewcommand{\arraystretch}{1.}
    \begin{tabular}{cccccccc}
    
    GT &
     Stable Diff.&
     R-ESRGAN &
     HAT &
     LIIF &
    Interp. &
     Ours &
    \\
         \tikz{ 
        \node[draw=black, line width=.5mm, inner sep=0pt] 
        {\includegraphics[width=.14\linewidth]{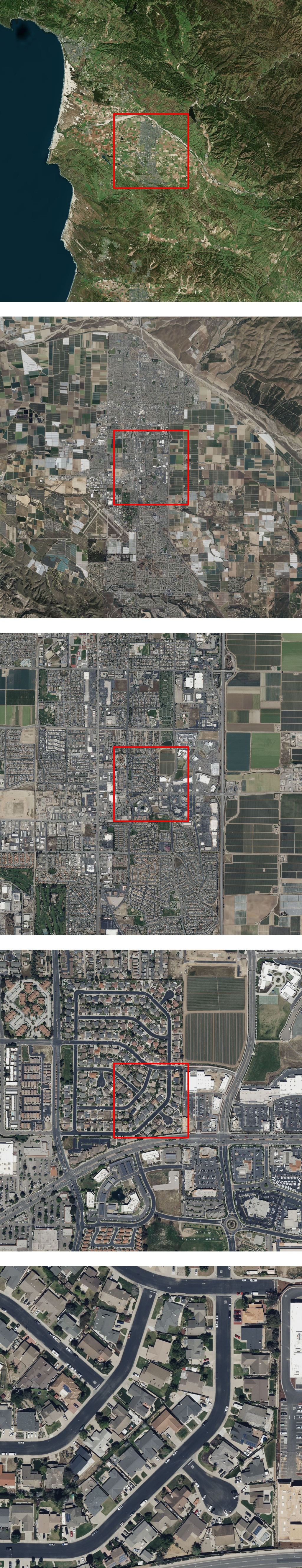}}
        }&
        \tikz{
        \node[draw=black, line width=.5mm, inner sep=0pt] 
        {\includegraphics[width=.14\linewidth]{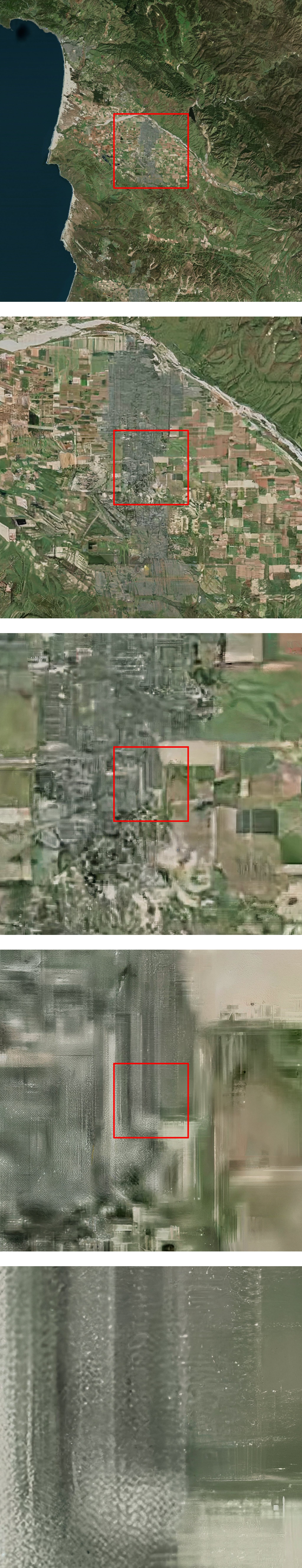}}
        }&
       
         \tikz{
        \node[draw=black, line width=.5mm, inner sep=0pt] 
        {\includegraphics[width=.14\linewidth]{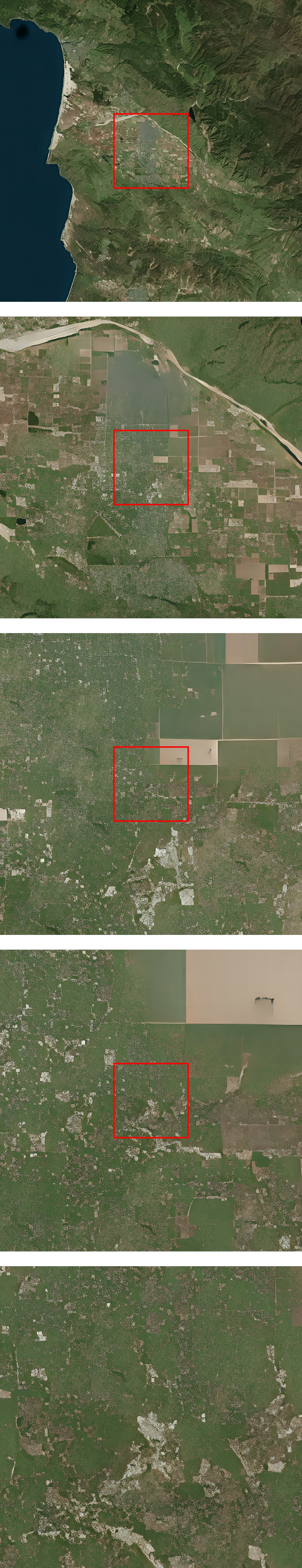}}
        }&
        
         \tikz{
        \node[draw=black, line width=.5mm, inner sep=0pt] 
        {\includegraphics[width=.14\linewidth]{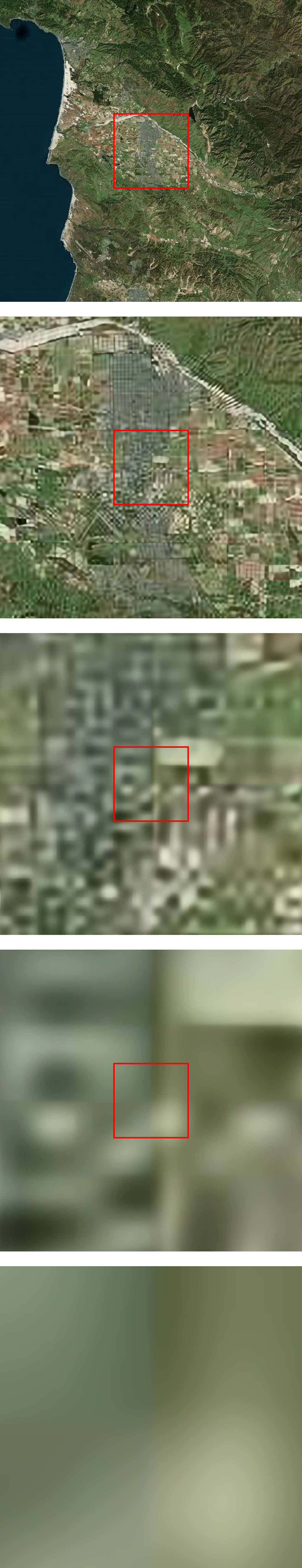}}
        }&
        \tikz{ 
        \node[draw=black, line width=.5mm, inner sep=0pt] 
        {\includegraphics[width=.14\linewidth]{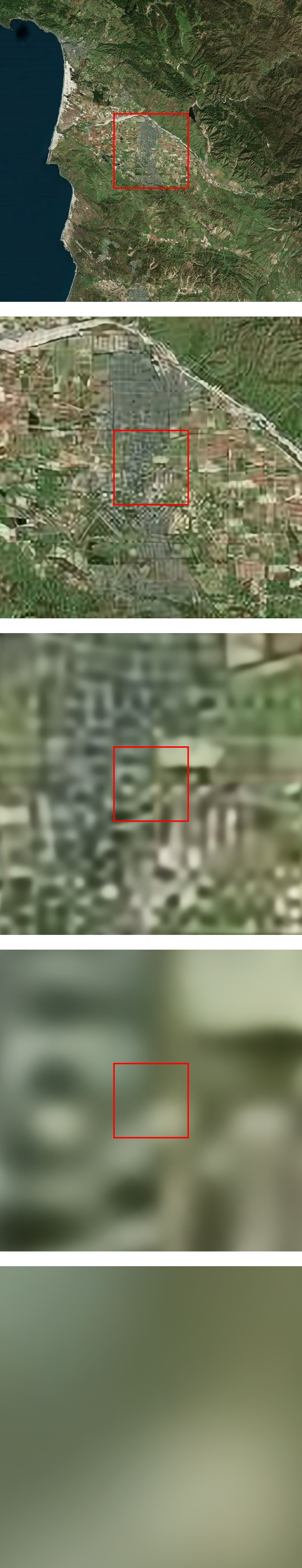}}
        }&
         \tikz{
        \node[draw=black, line width=.5mm, inner sep=0pt] 
        {\includegraphics[width=.14\linewidth]{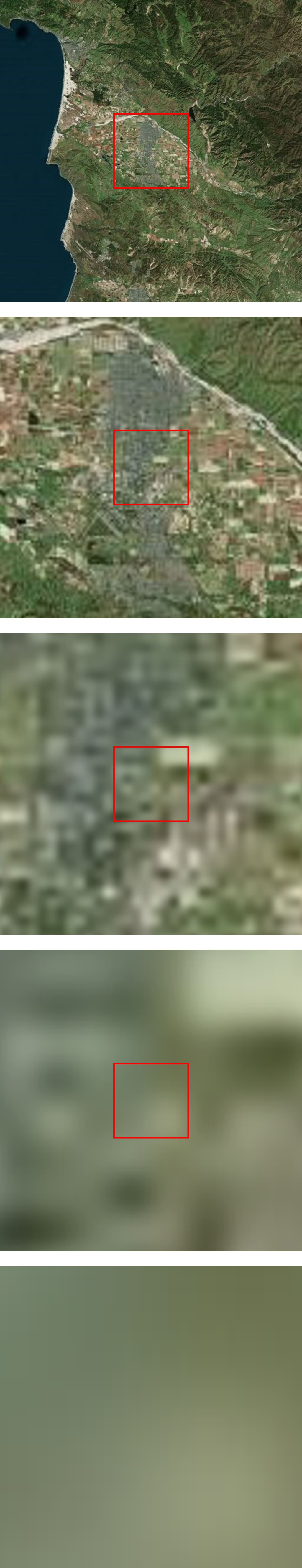}}
        }&

        \tikz{
        \node[draw=black, line width=.5mm, inner sep=0pt] 
        { \includegraphics[width=.14\linewidth]{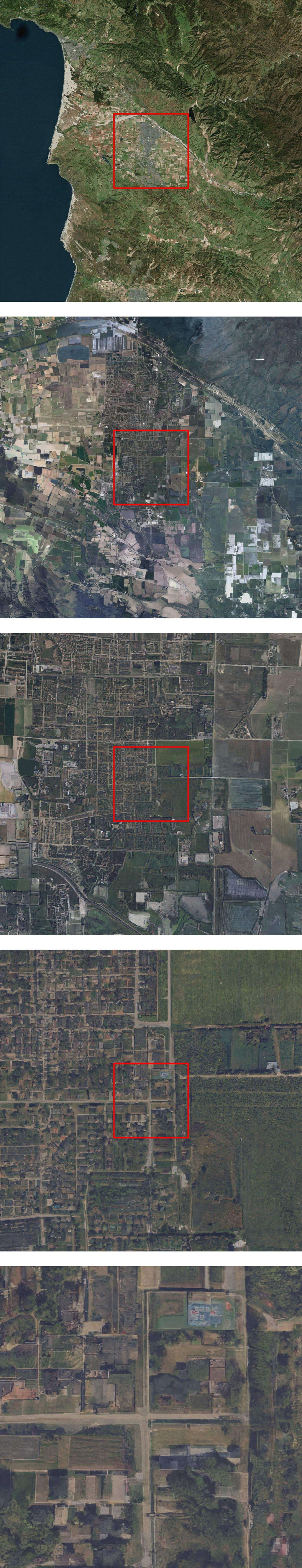}}
        }
    \end{tabular}
}
    \captionof{figure}{Qualitative Super-Resolution Comparison}
    \label{fig:qual_comp}
    \vspace{-30pt}
\end{table*}

\subsection{Applications}
\label{sec:applications}

\begin{figure}[!htbp]
    \vspace{-5pt}
    \centering
    \includegraphics[width=\linewidth]{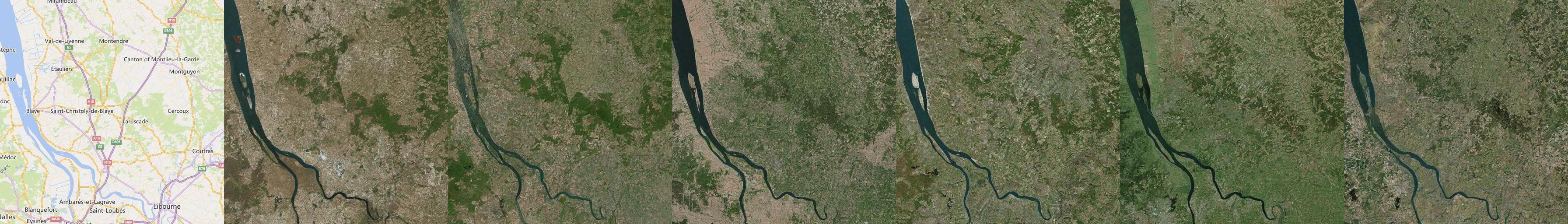}
    \vspace{0.5cm} 
    \includegraphics[width=\linewidth]{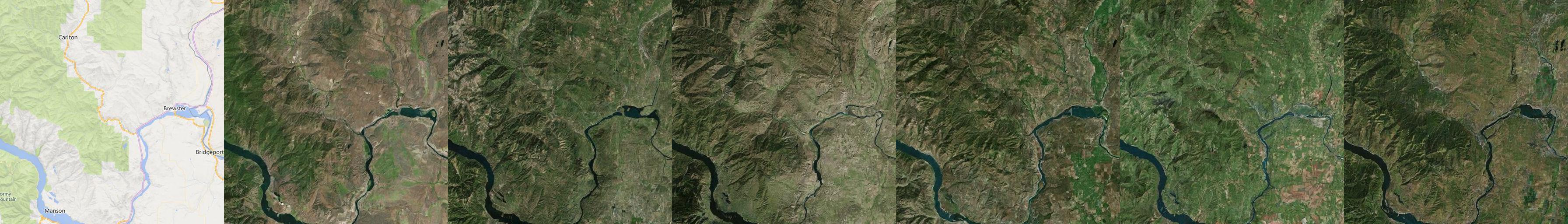}
    \vspace{-25pt}
    \caption{Sample Map-Conditioned Generations. Observe the model's ability to closely follow the map features while diversely filling in the details over different samples.}
    \label{fig:control}
    \vspace{-10pt}
\end{figure}

\paragraph{Controllable World Generation.}
Though EarthGen can be used to unconditionally generate outputs, we provide an example of base-layer conditioning to create controllable generations. In particular, we introduce map conditioning via augmenting the base layer model with a ControlNet~\cite{controlnet}, allowing the model to learn to generate map-consistent terrains with minimal changes. Fig. \ref{fig:control} demonstrates the potential of this expansion to enable controllable and diverse terrains. 

\paragraph{3D Scene Generation.}
    Off-the-shelf depth estimation methods such as DepthAnything \cite{depthanything} can be used to transform our 2D orthographic generations into 3D worlds. We do this by generating a triangle mesh from the 3D point cloud given by the RGB-D image. We include an example of such a 3D generation.

\begin{figure}[!tb]
    \centering
    \includegraphics[width=0.49\linewidth]{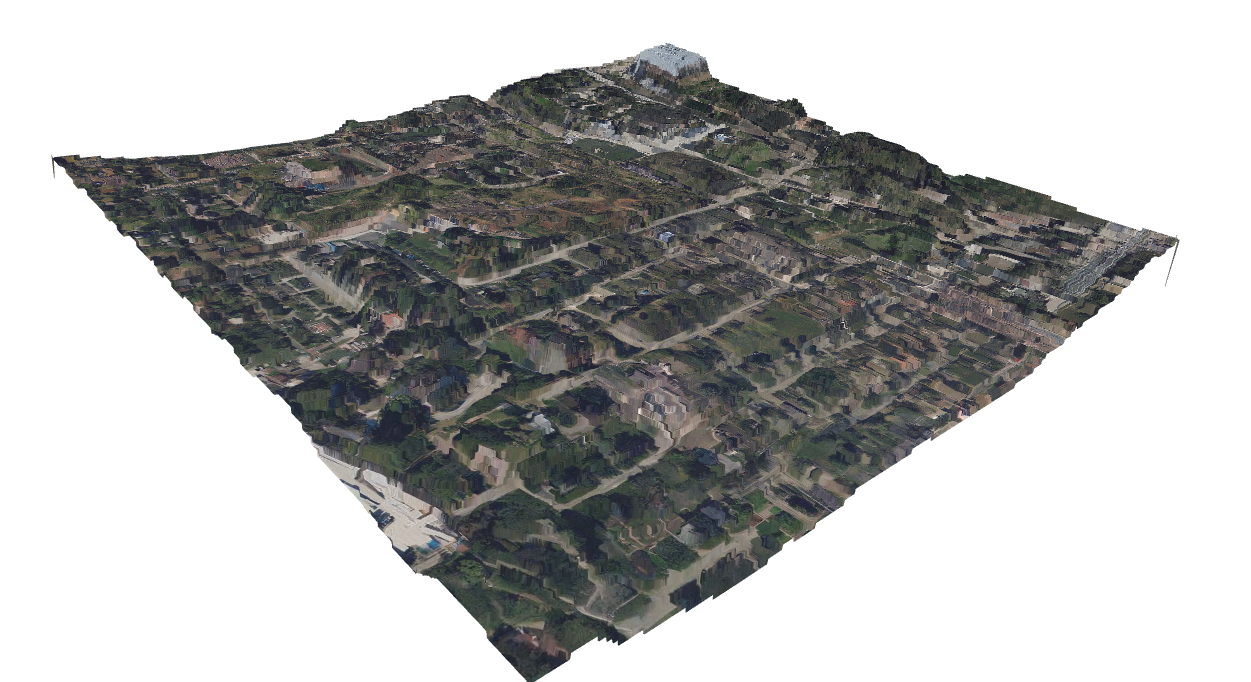}
    \hfill
    \includegraphics[width=0.49\linewidth]{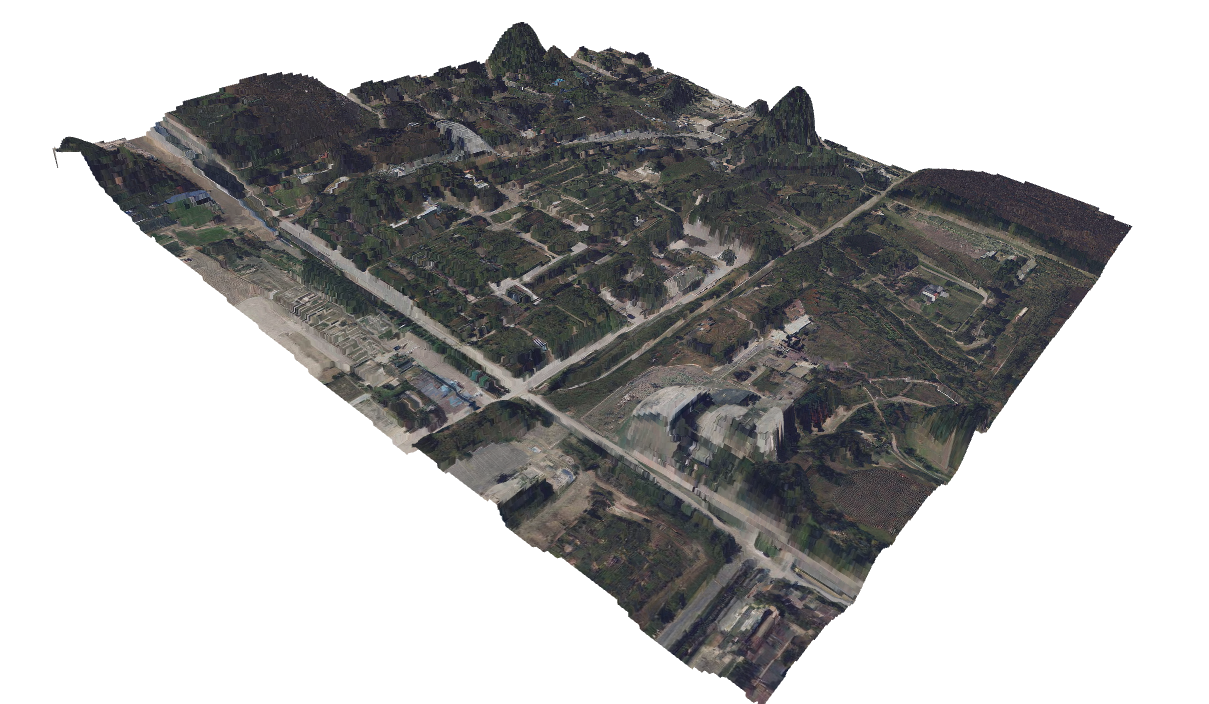}
    \caption{3D triangle meshes generated from EarthGen outputs using DepthAnything \cite{depthanything}. We can see clear road outlines and realistic elevations for trees and buildings.}
    \label{fig:depthanything}
\end{figure}

\subsection{Method Ablations}

\begin{figure}[!htbp]
    \centering
    \begin{subfigure}{0.49\textwidth}
        \centering
        \resizebox{\linewidth}{!}{
            \setlength{\tabcolsep}{0.2em}
            \renewcommand{\arraystretch}{1.}
            \begin{tabular}{cc}
                Naive Stitching & Gaussian Comp. \\
                \tikz{\node[draw=black, line width=.5mm, inner sep=0pt]{\includegraphics[width=.49\linewidth]{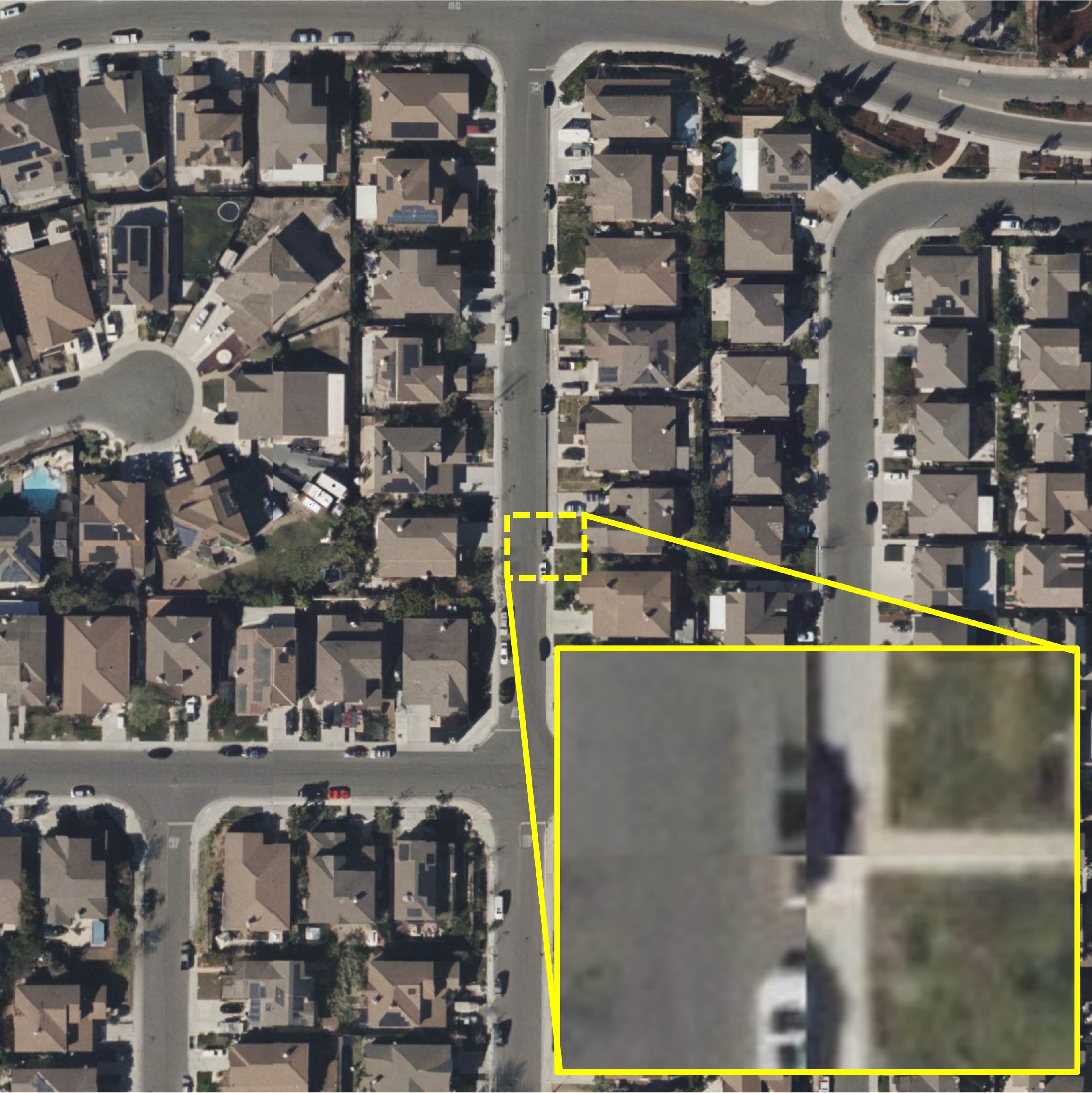}}} &
                \tikz{\node[draw=black, line width=.5mm, inner sep=0pt]{\includegraphics[width=.49\linewidth]{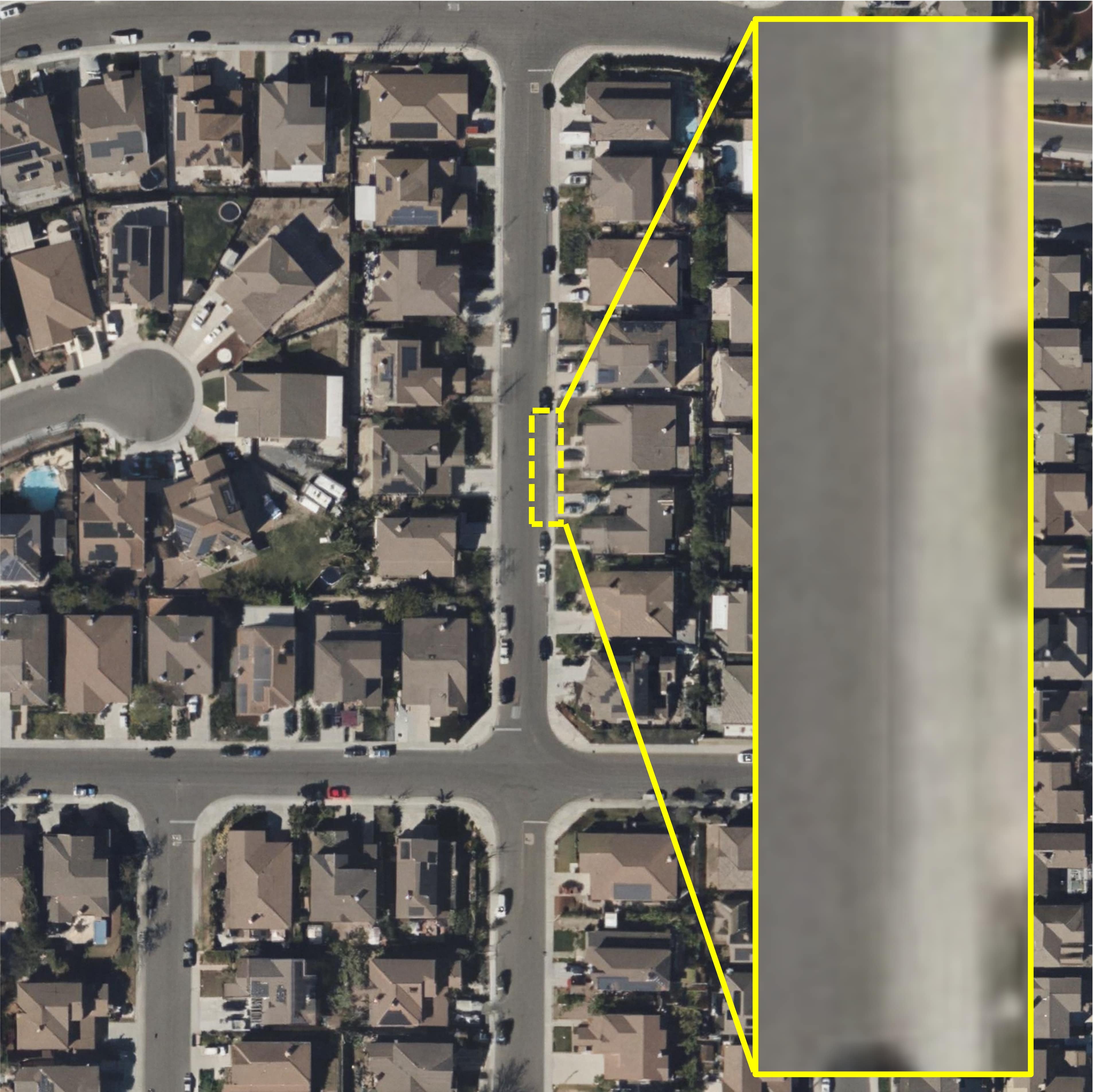}}} \\
                MultiDiffusion & Mix. of Diffusers \\
                \tikz{\node[draw=black, line width=.5mm, inner sep=0pt]{\includegraphics[width=.49\linewidth]{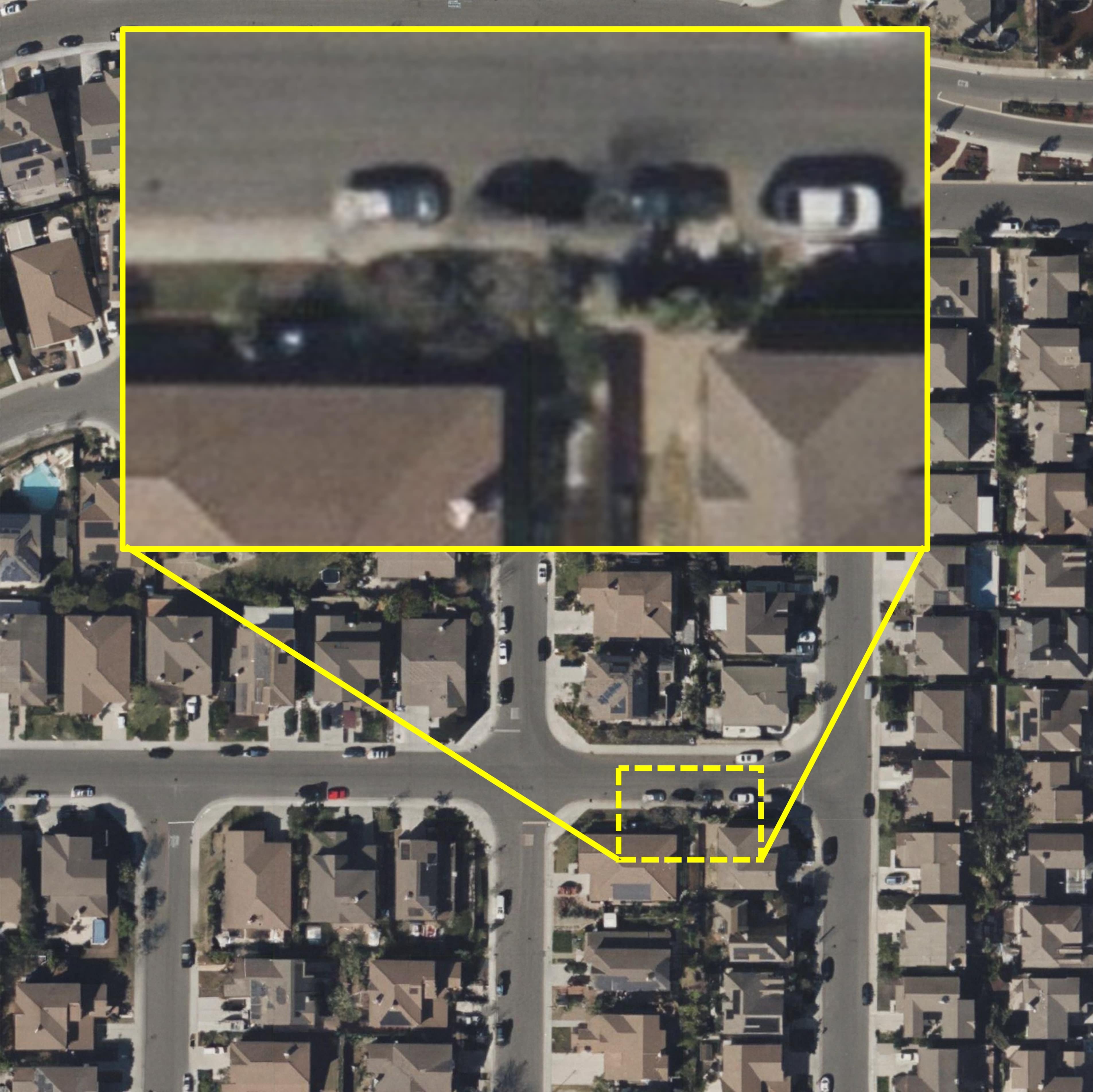}}} &
                \tikz{\node[draw=black, line width=.5mm, inner sep=0pt]{\includegraphics[width=.49\linewidth]{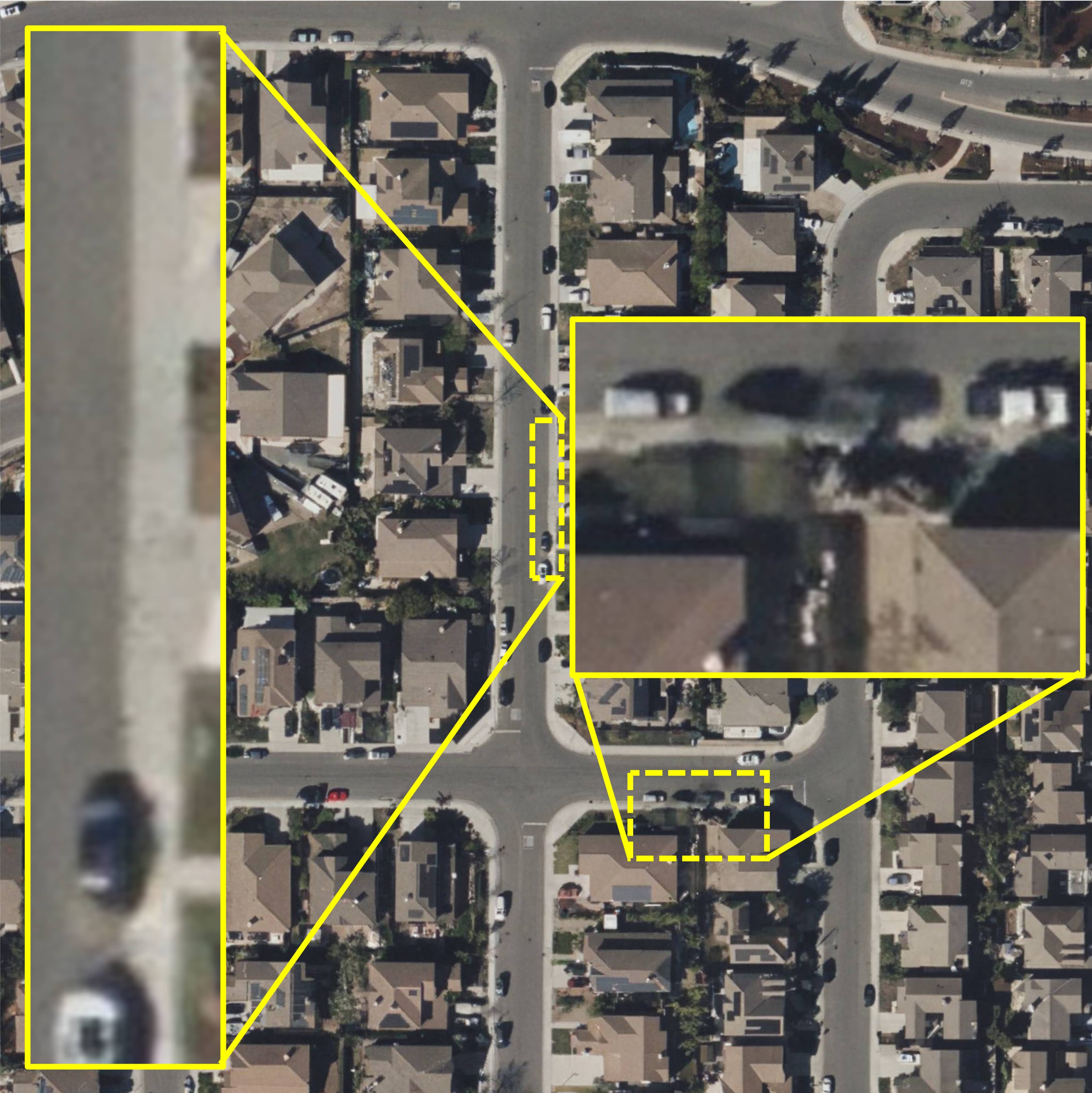}}}
            \end{tabular}
        }
        \caption{}
        \label{fig:tiling_abl}
    \end{subfigure}
    \hfill
    \begin{subfigure}{0.49\textwidth}
        \centering
        \resizebox{\linewidth}{!}{
            \setlength{\tabcolsep}{0.2em}
            \renewcommand{\arraystretch}{1.}
            \begin{tabular}{cc}
                Cascaded & Direct  \\
                \tikz{\node[draw=black, line width=.5mm, inner sep=0pt]{\includegraphics[width=.49\linewidth]{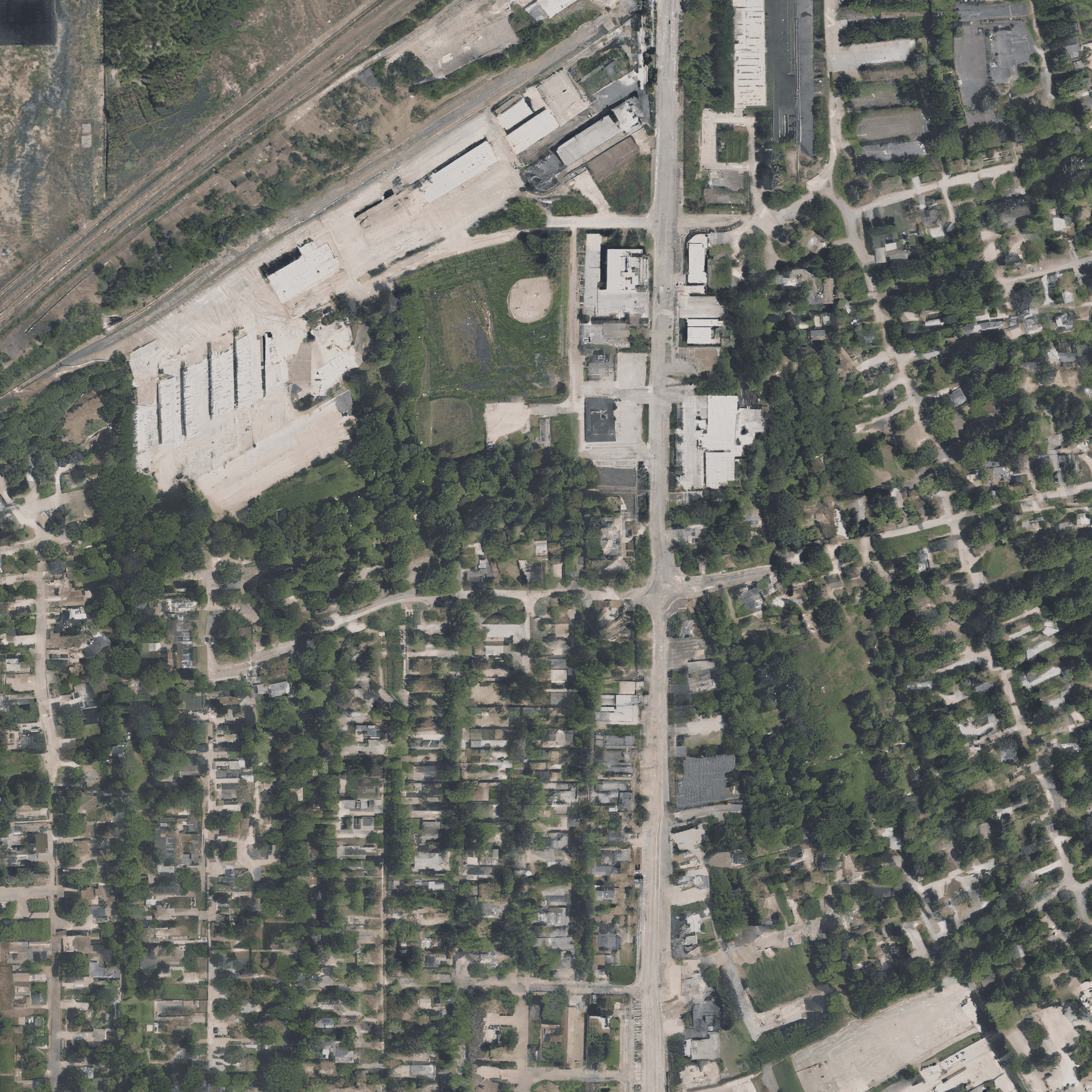}}} &
                \tikz{\node[draw=black, line width=.5mm, inner sep=0pt]{\includegraphics[width=.49\linewidth]{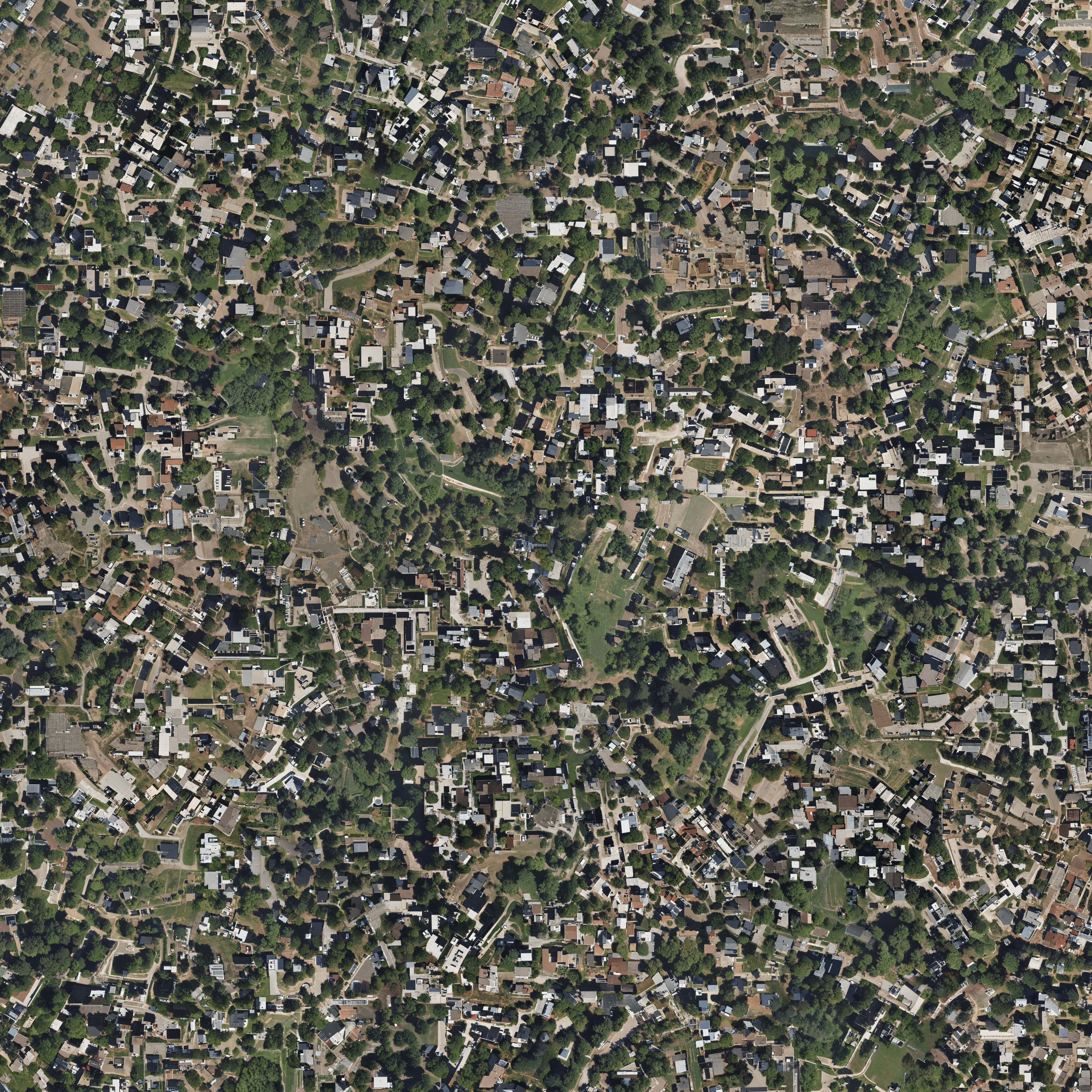}}} \\
                & \\
                \tikz{\node[draw=black, line width=.5mm, inner sep=0pt]{\includegraphics[width=.49\linewidth]{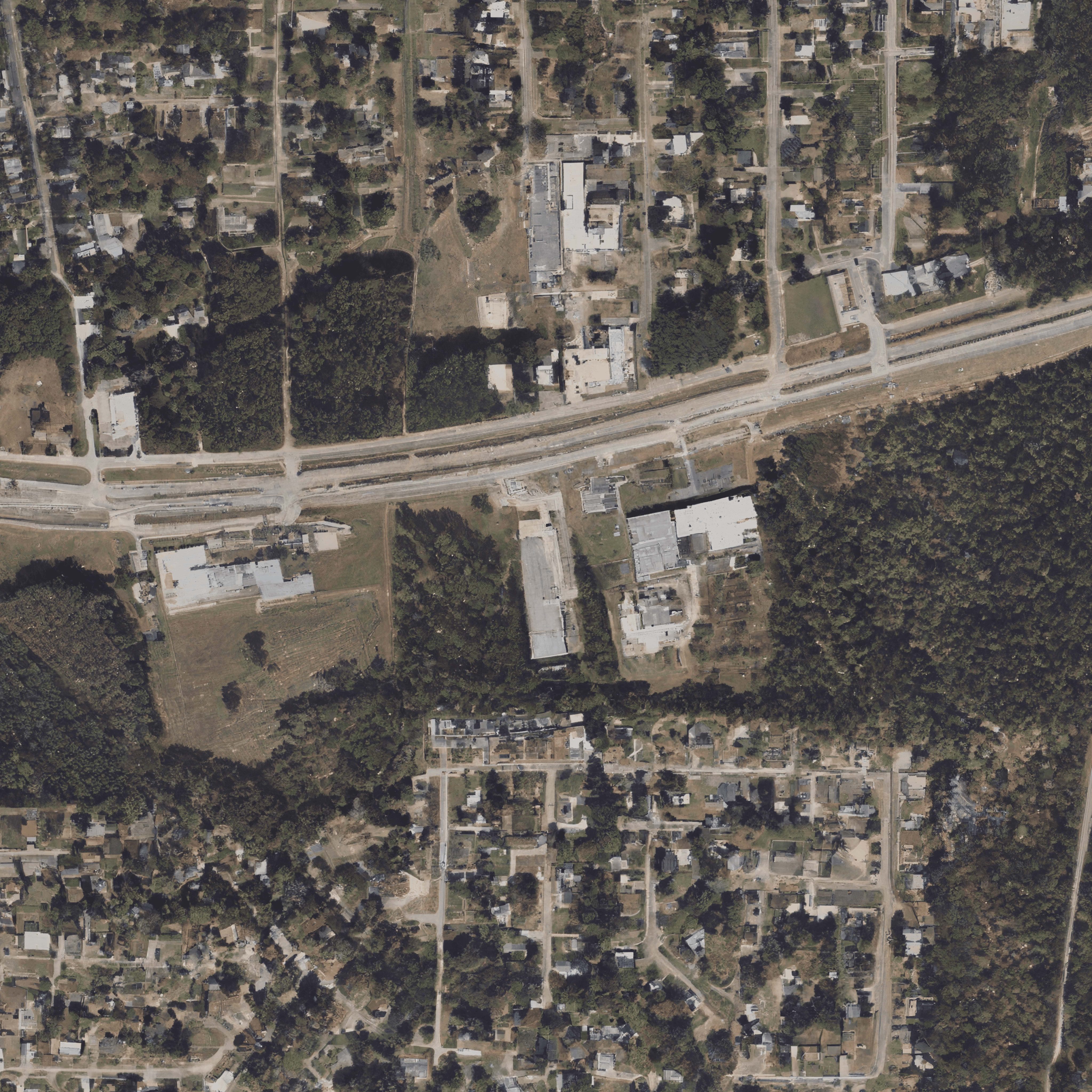}}} &
                \tikz{\node[draw=black, line width=.5mm, inner sep=0pt]{\includegraphics[width=.49\linewidth]{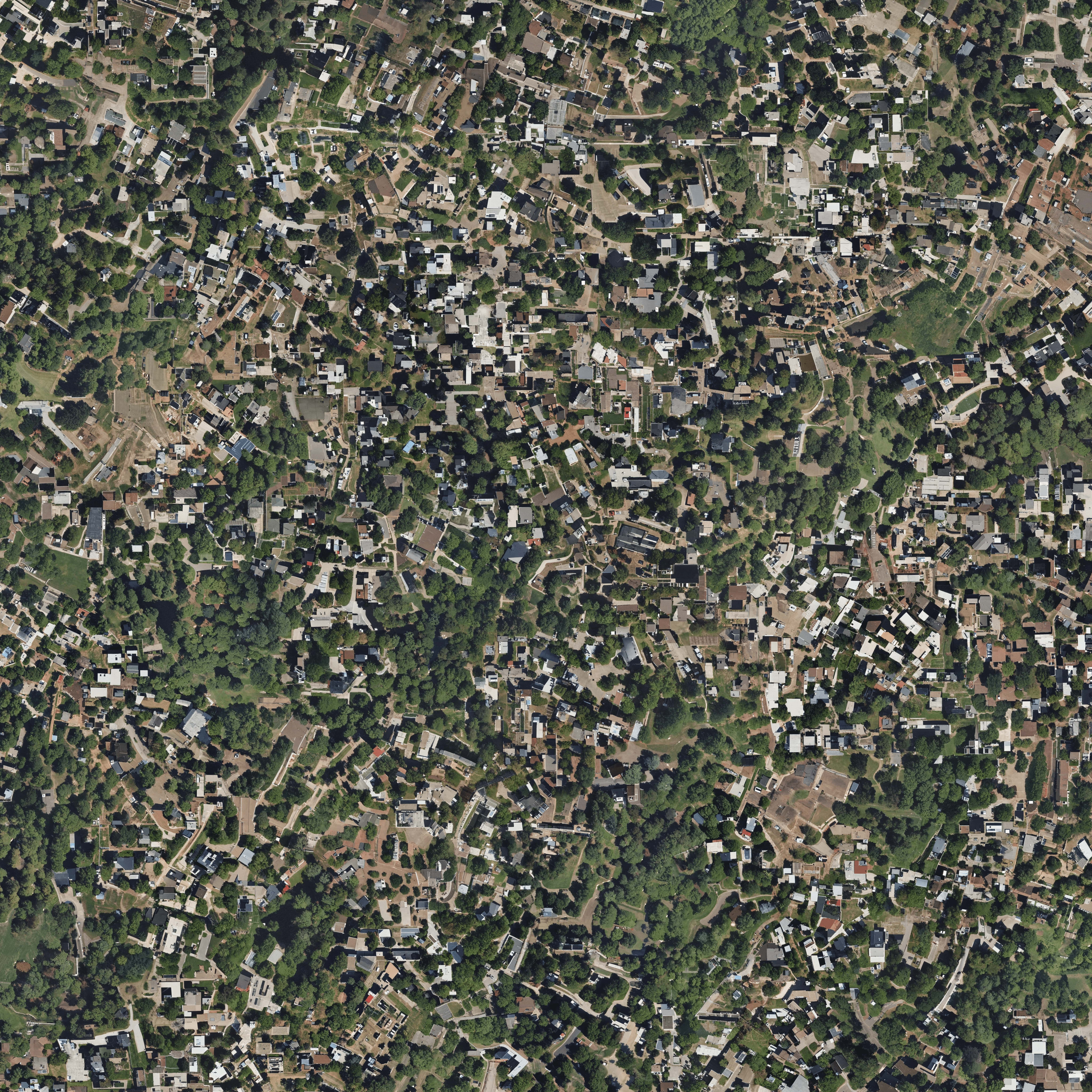}}}
            \end{tabular}
        }
        \label{fig:cascade}
        \caption{}
    \end{subfigure}
    \caption{\textbf{(a)} Tiling Ablations. Comparison of different tiling methods. Here, we see seams from all methods except Mixture of Diffusers. \textbf{(b)} Cascaded vs Direct Unconditional Generation. Direct Generation produces an unrealistic global structure despite looking good locally. Cascaded generation generates plausible structure both locally and globally.}
    \label{fig:ablations}
\end{figure}

\paragraph{Tiling Methods. } We explored four different tiling methods, including naive stitching, Gaussian compositing of overlapping tiles, Multidiffusion~\cite{bar2023multidiffusion}, and Mixture of Diffusers~\cite{jimenez2023mixture}. We found that using Mixture of Diffusers greatly improves the quality of generated images compared to other methods. To demonstrate this, we use each tiling method during one step of superresolution and qualitatively compare the artifacts from each tiling method in Fig. \ref{fig:ablations} (a). 

We observe that stitching produced visible seams between tiles even at one zoom. For Gaussian compositing, seams are slightly better hidden, but significant blurring is introduced. Multidiffusion is a large improvement from the other two but still produces lightly visible seams, the most clear one in Fig. \ref{fig:ablations} (a) being a horizontal line towards the right of the image in the grass. Mixture of Diffusers produces no artifacts in areas that other methods did, and tiling artifacts are generally nonexistent.

\paragraph{Direct Generation. } Using a cascaded pipeline is crucial to the structure of the generated large images. We explore an ablation by naively generating a larger scale terrain directly at the highest resolution in order to highlight this. We can qualitatively see the effects of the cascaded generation by comparing the outputs of our multi-layer cascade generation with those of a tuned, unconditional baseline designed to solely generate at the highest resolution in Figure \ref{fig:ablations} (b). 

We observe that cascaded generation is highly effective in producing a large, realistic image, whereas an unconditionally generated baseline only looks realistic locally. Although the unconditional baseline generates tiles of neighborhoods that look good when zoomed in, these tiles make no structural sense when viewed as a whole. Furthermore, features larger than a single tile such as roads and large buildings are only possible with the cascade introducing them at earlier stages.

\section{Discussion}
\setlength{\parindent}{15pt}
EarthGen marks a significant advancement in the field of large-scale terrain generation. By combining the strengths of hierarchical and compositional generation methods, our framework is capable of producing highly realistic and infinitely scalable terrain images. The generated terrains exhibit an unprecedented level of detail and diversity, capturing the essence of Earth's varied natural and man-made landscapes. There remain some limitations due to the dataset which are discussed further in the supplementary material. 

We note that EarthGen can be easily combined with downstream pipelines to enable countless applications, many with the potential to help address upcoming environmental and societal challenges. Immediate use cases in environmental monitoring and agricultural resource management can be attained by using EarthGen to augment remote sensing datasets for tasks such as land cover classification and automatic mapping. Conditional content creation is easily enabled in a similar manner as we demonstrated with ControlNet, allowing for asset creation for open-world 2D games or even 3D games if paired with a depth module such as DepthAnything. Future work will explore pairing EarthGen with an action conditioning model, allowing for visualizing/evaluating urban planning decisions before implementing for more informed decision making. We open source our code and models at \url{https://github.com/anshgs/earthgen} in hopes of accelerating research to realize these benefits.

\bibliographystyle{splncs04}
\bibliography{main}
\setcounter{section}{0}
\renewcommand*{\thesection}{\Alph{section}}
\setlength{\parindent}{15pt}

\title{Supplementary Material\newline Generating the World from Top-Down Views}
\titlerunning{Supplementary Material: Generating the World from Top-Down Views}
\author{}
\institute{}
\maketitle

Sec.~\ref{sec:datadetails} provides additional implementation details about dataset considerations. Sec.~\ref{sec:derivations} contains derivations for equations from the main paper. Sec.~\ref{sec:further} describes further experiments and ablations. Sec.~\ref{sec:vis} provides extra visualizations for baseline comparisons as well as discusses failure cases for our method.

\setcounter{section}{0}
\renewcommand*{\thesection}{\Alph{section}}
\setlength{\parindent}{15pt}

\section{Dataset Details}
\label{sec:datadetails}

\begin{figure}[!htbp]
    \centering
    \includegraphics[scale=0.45]{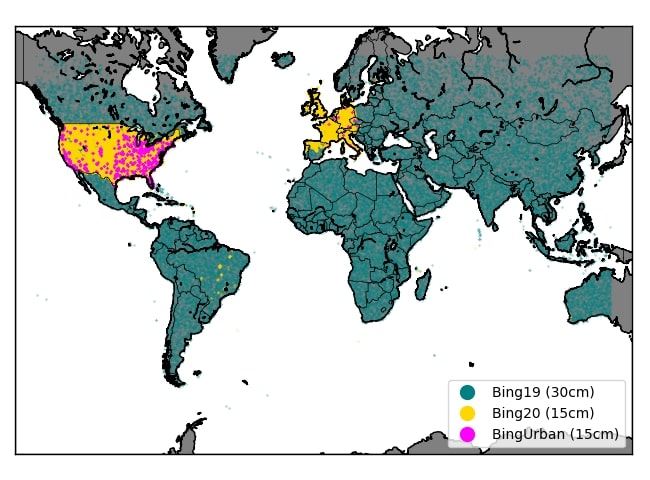}
    \caption{Dataset Distributions: Zoom 20 data (15 cm/px) is mostly available only over America and Western Europe while Zoom 19 (30 cm/px) is available over the rest of the land in the world. High-resolution ocean tiles are typically unavailable.}
    \label{fig:dataset}
\end{figure}

Our dataset is collected in three portions from Bing Maps~\cite{bingmaps}. Bing19 is collected with image resolution up to 30 cm/px, while Bing20 and BingUrban are collected up to 15 cm/px. We note a few key considerations to be aware of with using datasets compiled from similar sources. 

Firstly, the distribution of resolutions in this dataset is imbalanced - the highest resolution is only available in Western Europe and the United States. This may contribute in part to our models being biased towards generating Western architecture and terrains as opposed to matching the actual world distribution. 

A more intrinsic issue with the dataset is the sharp and conspicuous color shift between Bing's zoom 12 and zoom 13, which is likely caused by the different zoom levels being aggregated from different data sources and/or imaging equipment. This leads to inconsistencies with the ground truth between those layers. Although this shift is learnable to some degree, this introduces the potential for significant hallucination due to the sharp jump in detail over that zoom change. This color shift is quite visible in the ground truth visualization in Fig. \ref{fig:qual_comp1}, \ref{fig:qual_comp2}, and \ref{fig:qual_comp3}, between the first image at zoom 12 and the second at zoom 14. However, as can be seen in our baseline comparison figures, our 12to14 model effectively learns this color shift.

Additionally, due to Bing Maps' possible data collection from different sources, we see seasonal effects in some locations, including snow, or autumn leaves, as well as other artifacts like clouds and cloud shadows. The aforementioned features may or may not be consistent with other zooms. As such, we see failure cases when the diffusion process diverges in cloud or snow cases.

Finally, our construction of these datasets heavily undersamples water tiles. Ocean tiles typically only had resolution up to 19 m/px, so we entirely avoided them while constructing our datasets. Combined with the detail shift between zoom 12 and 14, this undersampling commonly leads to divergence in water cases during the diffusion process, especially when text conditioning is added on as highlighted in Fig. \ref{fig:fail}.

\red{
\section{Negative Conditioning Derivations}
\label{sec:derivations}
Following the formulation from \cite{liu2022compositional}, we can derive Eq. 4 and Eq. 5 as follows: 

$$p_\theta(\bx_s| \neg \mathbf{e}^-, \bx_{s-1}) \propto p_\theta(\bx_s,  \neg \mathbf{e}^- | \bx_{s-1}) \propto \frac{p_\theta(\bx_s| \bx_{s-1})}{p_\theta(\mathbf{e}^{-} | \bx_s, \bx_{s-1})}$$

Using the factorization $p(c_j | x) \propto \frac{p(x|c_j)}{p(x)}$, we get to our desired expression for Eq. 4:

$$p(\bx_s|\neg \mathbf{e}^{-}, \bx_{s-1}) \propto \frac{p_\theta(\bx_{s} | \bx_{s-1})^2}{p(\bx_s|\mathbf{e}^{-}, \bx_{s-1})}$$

Noting that the score function can be interpreted as the gradient of the negative log likelihood of the probabilities, we can then express this as a composed score function by introducing a weighting parameter $\lambda$ to control the strength of the negative parameter while maintaining the expected mean of the predicted noise:
$$\hat{\epsilon}(x_t, t) = \epsilon_\theta(x_t, t) + \lambda (\epsilon_\theta(x_t, t) - \epsilon_\theta (x_t, t | c_j))$$

}
\section{Additional Experiments}
\label{sec:further}

\subsection{Negative Text Guidance Ablation} Using negative text conditioning is key to achieving realistic and high-quality images. To demonstrate the effectiveness of negative text conditioning, we ran inference on our validation set with and without negative text guidance. Quantitatively, as shown in Table \ref{tab:ablation} adding negative text ablation outperforms the lack thereof in both FID and KID. Qualitatively, we can see from Fig. \ref{fig:neg_text_abl} that adding negative text conditioning results in more realistic and better-quality images, especially at the final layer. This is reflected in improvements in both urban and non-urban terrains, as the textures and structures become more realistic with the negative prompting as compared to without. 

\begin{table}[!htbp]
  \centering
  \begin{tabular}{c|cccc}
    Negative Conditioning &  FID($\downarrow$) & KID ($\downarrow$) \\
    \midrule
    No & 87.03/171.67 & 0.0549/0.1474\\
    Yes & \textbf{66.40/91.04} & \textbf{0.0210/0.0533} \\
    \bottomrule
  \end{tabular}
  \vspace{5mm}
  \captionof{table}{General/Urban Metrics for Negative Guidance Ablation. Negative Conditioning improves quantitative performance across datasets/metrics}
  \label{tab:ablation}
\end{table}

\begin{table*}[]
    \label{fig:textabl} 
    \centering
    \resizebox{\linewidth}{!}{
\setlength{\tabcolsep}{0.2em} %
\renewcommand{\arraystretch}{1.}
    \begin{tabular}{cc}
    
    \multicolumn{2}{c}{\,\,\,\,\,\,\,\,\,\,\,\,\,\,\,With Text Conditioning} \\
    Urban\vspace{1mm} & \raisebox{-0.5\height}{\tikz{ 
        \node[draw=black, line width=.5mm, inner sep=0pt] 
        {\includegraphics[angle=90,width=.8\linewidth]{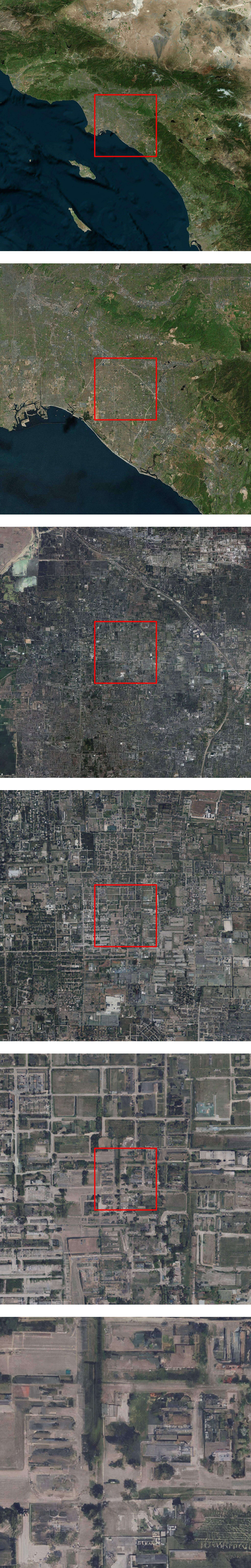}}
        }} \\
    20\vspace{1mm} & \raisebox{-0.5\height}{\tikz{ 
        \node[draw=black, line width=.5mm, inner sep=0pt] 
        {\includegraphics[angle=90,width=.8\linewidth]{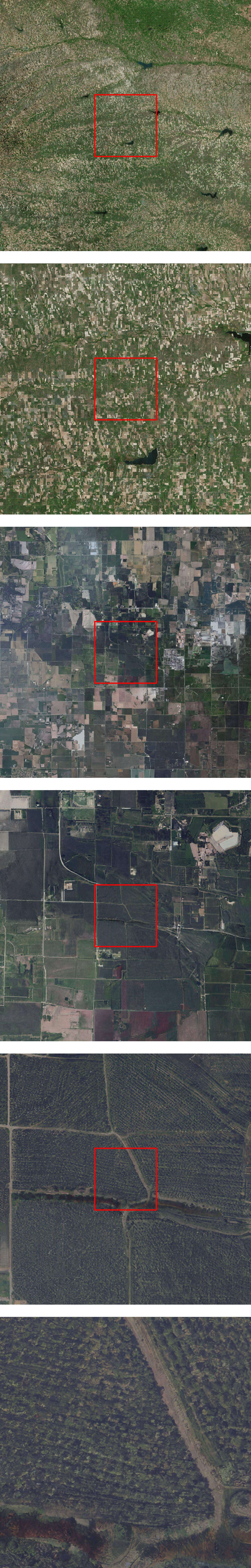}}
        }} \\
    \multicolumn{2}{c}{\,\,\,\,\,\,\,\,\,\,\,\,\,\,\,Without Text Conditioning} \\
     Urban\vspace{1mm} & \raisebox{-0.5\height}{\tikz{ 
        \node[draw=black, line width=.5mm, inner sep=0pt] 
        {\includegraphics[angle=90,width=.8\linewidth]{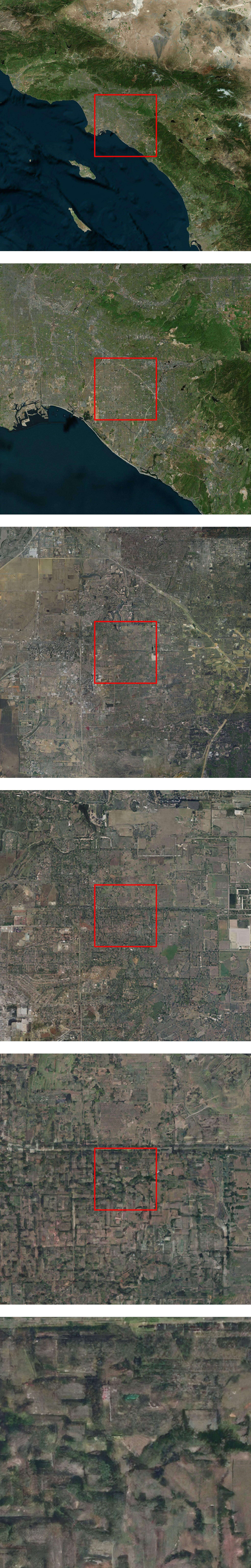}}
        }} \\
    20\vspace{1mm} & \raisebox{-0.5\height}{\tikz{ 
        \node[draw=black, line width=.5mm, inner sep=0pt] 
        {\includegraphics[angle=90,width=.8\linewidth]{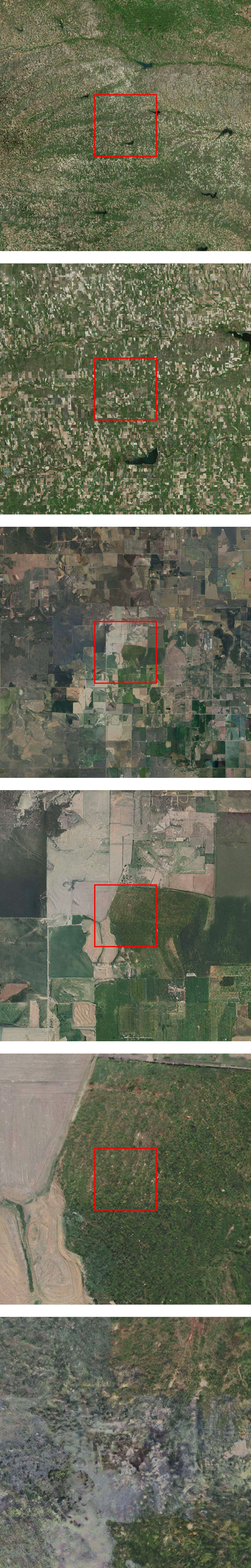}}
        }}

    \end{tabular}
    }
      \vspace{5mm}
    \captionof{figure}{Adding negative text conditioning produces more realistic, better-quality images, with recognizable structure and texture. Without the negative text conditioning, errors compound between layers much more significantly, preventing well-defined structures and textures from forming and instead only prioritizing consistency with potentially flawed prior generations. }
    \label{fig:neg_text_abl}
\end{table*}

\subsection{Remote Sensing Segmentation} We use an off-the-shelf segmentation model for remote sensing images~\cite{senseearth2020} to validate our realistic generations quantitatively (Table~\ref{tab:seg_quant}) and qualitatively (Fig.~\ref{fig:seg_qual}) below. The model fine-tunes existing models~\cite{zhao2017pyramid, wang2020deep} on remote sensing images, using water, ground, low vegetation, trees, buildings, and playgrounds as segmentation labels. Following the same procedure as in Section 4, we provide FID and KID metrics for $\times$1024 cascaded zoom 20 segmentation. Qualitatively, we see that the off-the-shelf segmentation model is able to discern most of the generated buildings, roads, and trees, showing that our model generates realistic enough structures to be recognized by models not fine-tuned on our outputs. 

\begin{table}[!htbp]
  \centering
  \begin{tabular}{ccccc}
    \toprule
    Method & FID($\downarrow$) & KID ($\downarrow$) \\
    \midrule
    SD x4 Upscaler & 9.20~/~43.40 & 0.0071~/~0.0466   \\
    Real ESRGAN & 11.99~/~50.42 & 0.0091~/~0.0534   \\
    HAT & 58.76~/~112.38 & 0.0638~/~0.1373  \\
    LIIF & 58.53~/~114.15 & 0.0639~/~0.1374   \\
    Interpolation & 60.88~/~117.07 & 0.0650~/~0.1377   \\
    \midrule
    \textbf{Ours} & \textbf{4.14~/~21.25} & \textbf{0.0014~/~0.0192}\\
    \bottomrule
  \end{tabular}
    \vspace{5mm}
  \caption{General/Urban $\times$1024 Segmentation Metrics. Our method strongly outperforms all baselines in all metrics/datasets.}
  \label{tab:seg_quant}
\end{table}

\begin{table}[]
    \centering
    \resizebox{\linewidth}{!}{
\setlength{\tabcolsep}{0.2em} %
\renewcommand{\arraystretch}{1.}
    \begin{tabular}{ccc}
    Layer 20 Generation &
    Corresponding Segmentation &
    \\
         \tikz{\node[draw=black, line width=.5mm, inner sep=0pt] {\includegraphics[width=.45\linewidth]{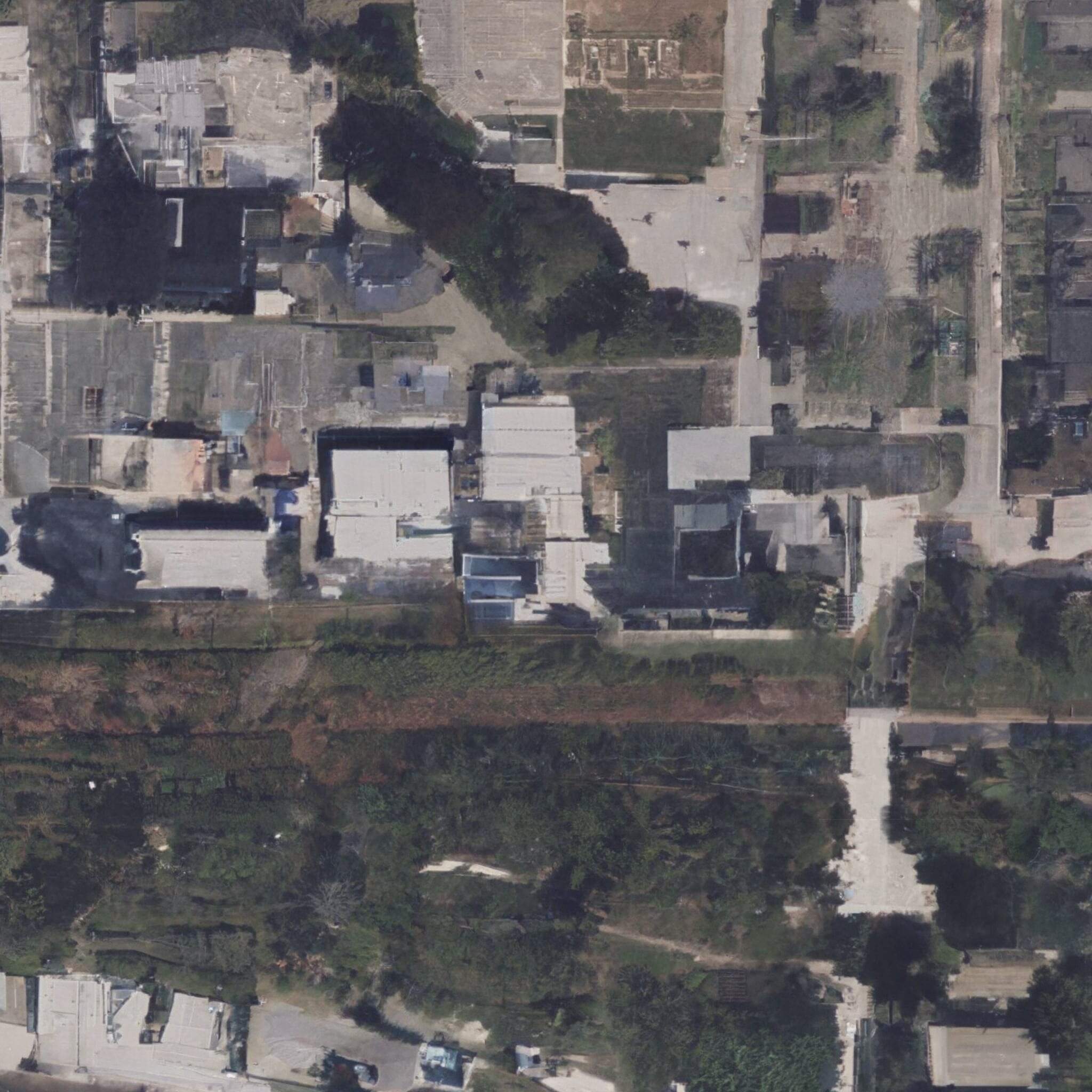}}}&\tikz{\node[draw=black, line width=.5mm, inner sep=0pt] { \includegraphics[width=.45\linewidth]{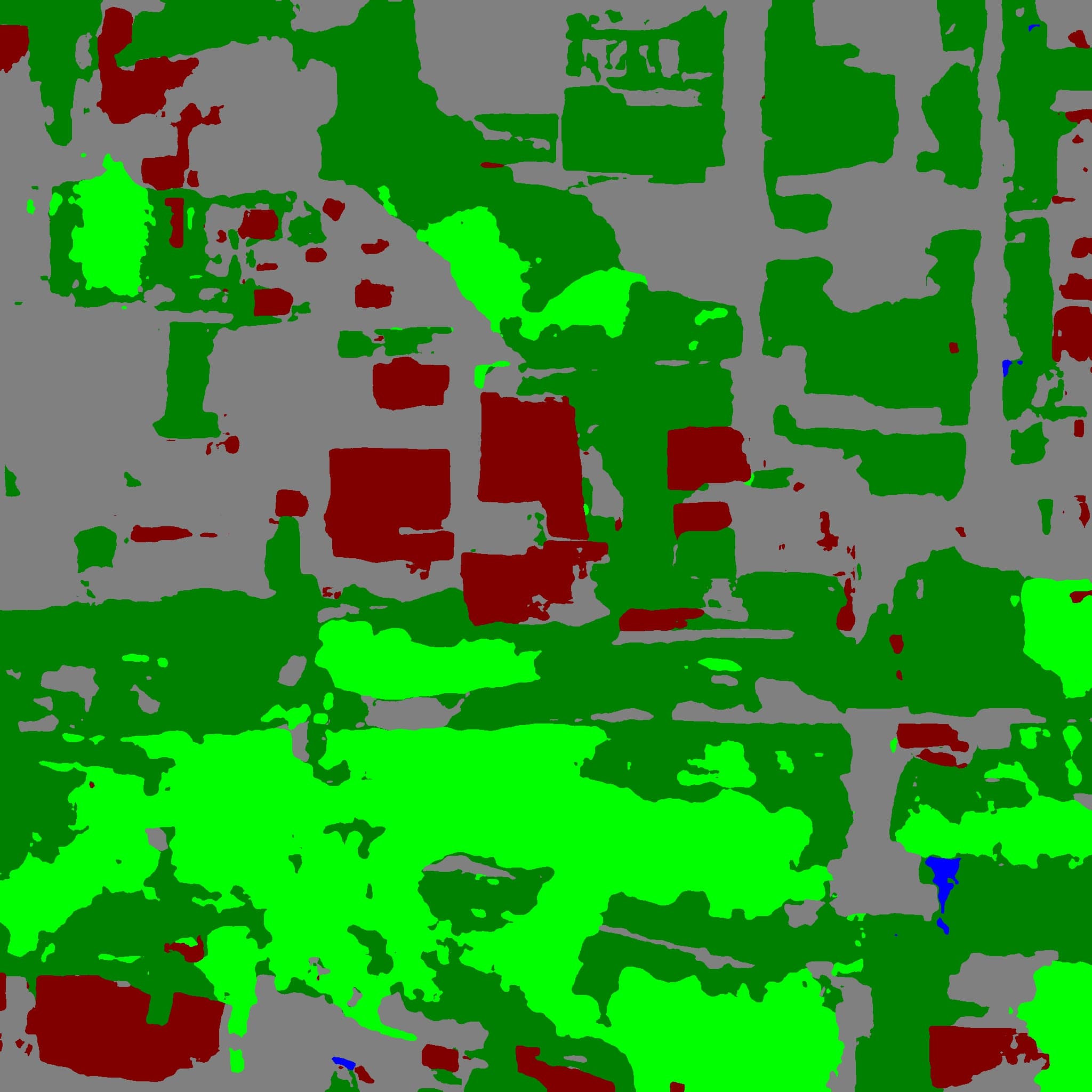}}}
    \end{tabular}
    }
    \captionof{figure}{ This generation yields realistic ``ground''(gray), ``low vegetation''(dark green), ``trees''(light green), and ``buildings''(red), as shown in the segmentation map.}
    \label{fig:seg_qual} 
\end{table}

\subsection{Text-Conditioned Generation}

\begin{figure}[!htbp]
  \centering
    \includegraphics[width=0.475\textwidth]{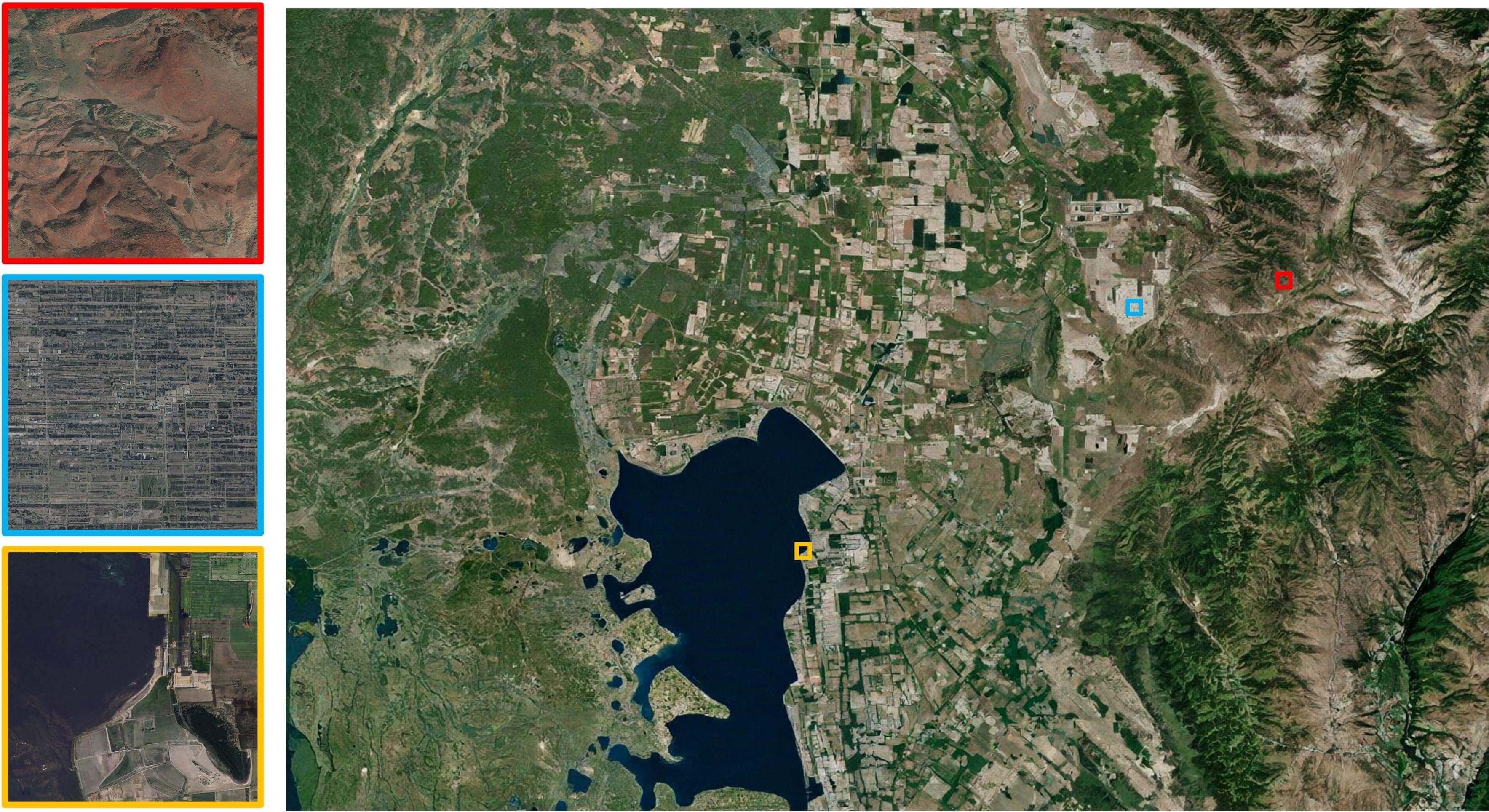}
  \caption{Text conditioned base layer generation for input labels of "lake", "city", and "mountains" from left to right.}
  \label{fig:condition}
\end{figure}

\begin{figure}[!htbp]
    \centering
    \includegraphics[scale=0.15]{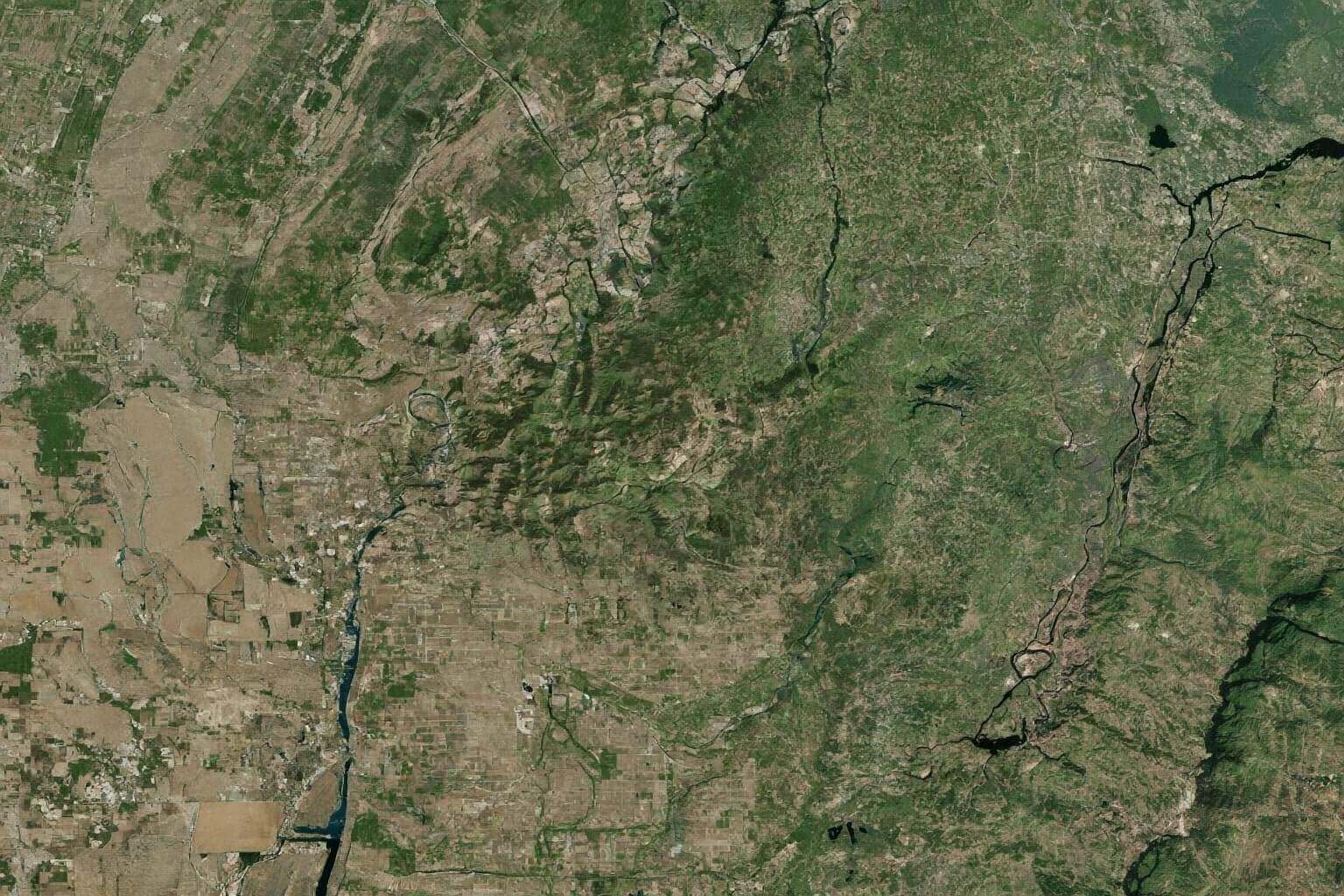}
    \caption{Base layer custom generation with ``town'' in left and bottom middle, ``hillside'' and ``river'' in top middle and right, and ``river'' in bottom right.}
    \label{fig:custom_2}
    \vspace{5pt}
    \includegraphics[scale=0.15]{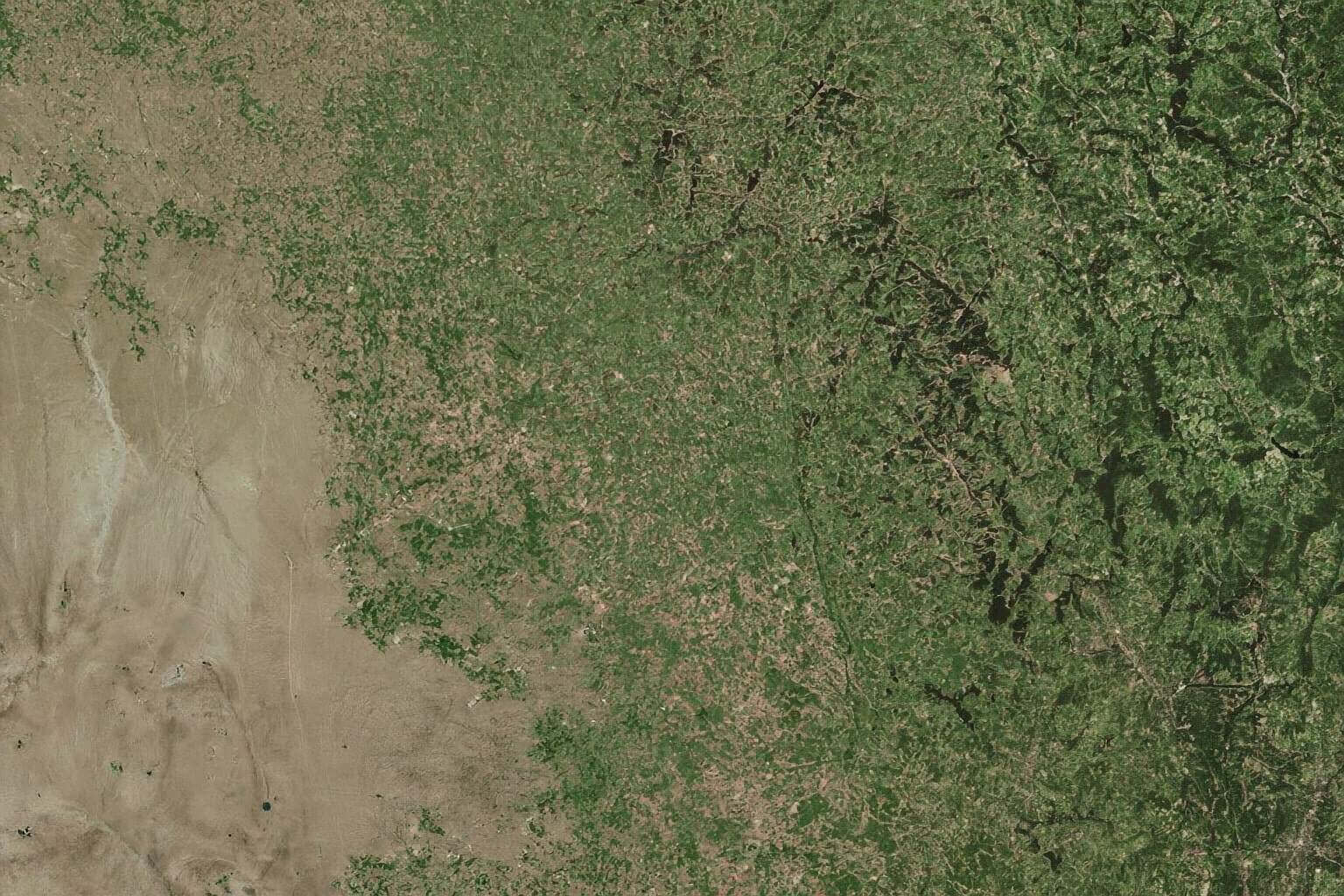}
    \caption{Base layer custom generation with ``dirt, grass'' in top left, ``dirt'' in bottom left, ``grass'' in middle, and ``forest'' in right.}
    \label{fig:custom_3}
\end{figure}

Though map conditioning data is not too difficult to generate due to the lack of fine details required, we provide an alternate, slightly weaker method, for controllable world generation as well via text. We train a conditional base layer generation module to take in label descriptions (ex. ``mountains") and generate resembling satellite imagery. We assign these labels using CLIP ~\cite{radford2021learning} during training time, and incorporate them during inference with classifier-free guidance by manually assigning labels to each tile. We demonstrate in Fig. \ref{fig:condition} that our tiling approach can combine geographically associated labels coherently while maintaining high quality. Further visualizations are included in Figs. \ref{fig:custom_2} and \ref{fig:custom_3}.

In order to accomplish this, we train an unconditional base layer model as described in Sec. 3.2 with no labels for 44k steps, followed by classifier-free guidance training with a single label where we randomly drop the label 10\% of the time for 32k additional steps. 

The label is chosen by the highest CLIP similarity with ground truth, over a predetermined list of single-word labels representing the common caption descriptors over a random sample of the dataset. 
A closed vocabulary for these captions was generated using ViT-GPT2 ~\cite{vit_gpt}, giving us the list ['water', `dirt', `lush', `forest', `grass', `field', `hillside', `town', `lake', `mountains', `city', `river', `neighborhood'].

At inference, one can specify desired labels for each tile of the grid. Fig. ~\ref{fig:condition} had labels of ``lake", ``lake, city", ``city", ``city, mountains" and ``mountains" over 5 half-overlapping columns. Positive label and negative guidance weights of 10 and 3, respectively, were used for base generation. Additional custom generations with their labels are provided in Fig.~\ref{fig:custom_2} and ~\ref{fig:custom_3}. While the examples shown are $1024\times1536$, the approach generalizes to arbitrarily large and sophisticated generations.

\section{Visualizations}
\label{sec:vis}
\subsection{Qualitative Baseline Comparisons}
We provide additional visualizations (Fig.~\ref{fig:qual_comp1},~\ref{fig:qual_comp2},~\ref{fig:qual_comp3}) of zoom pipeline comparisons between our model and each baseline when generating diverse terrain and features such as forests, roads, and farms. Our model consistently generates the highest quality images at the finest levels, while retaining consistency with the scene context.

    \begin{table*}[]
    \label{fig:baselines1} 
    \centering
    \resizebox{\linewidth}{!}{
\setlength{\tabcolsep}{0.2em} %
\renewcommand{\arraystretch}{1.}
    \begin{tabular}{cccccccc}
    
    GT &
     Stable Diff. &
     R-ESRGAN &
     HAT &
     LIIF &
    Interp. &
     Ours &
    \\
         \tikz{ 
        \node[draw=black, line width=.5mm, inner sep=0pt] 
        {\includegraphics[width=.14\linewidth]{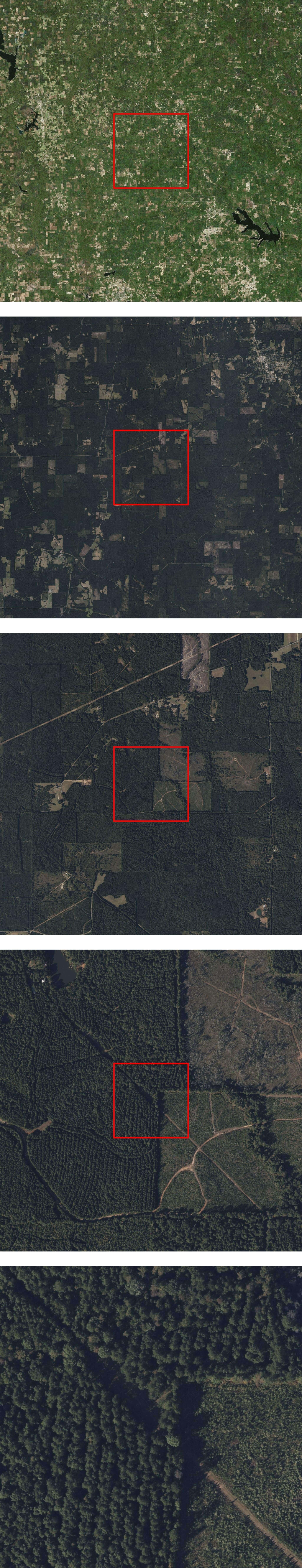}}
        }&
        \tikz{
        \node[draw=black, line width=.5mm, inner sep=0pt] 
        {\includegraphics[width=.14\linewidth]{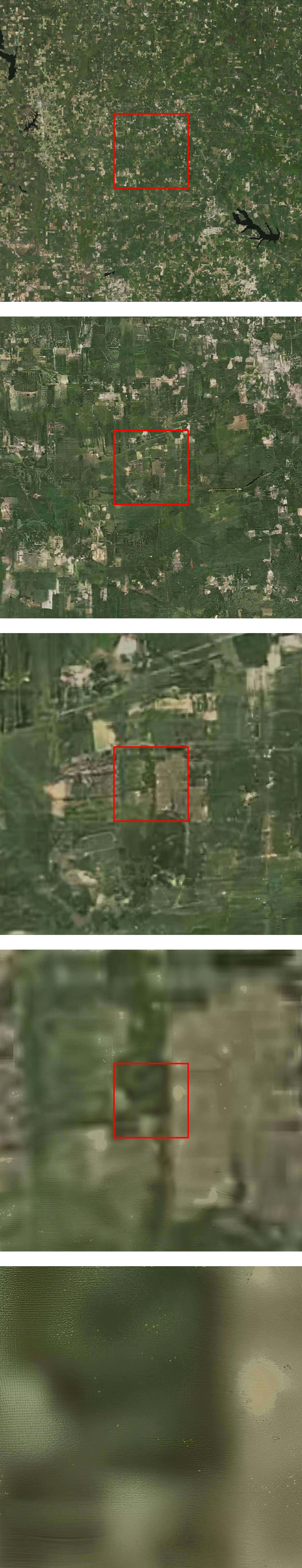}}
        }&
       
         \tikz{
        \node[draw=black, line width=.5mm, inner sep=0pt] 
        {\includegraphics[width=.14\linewidth]{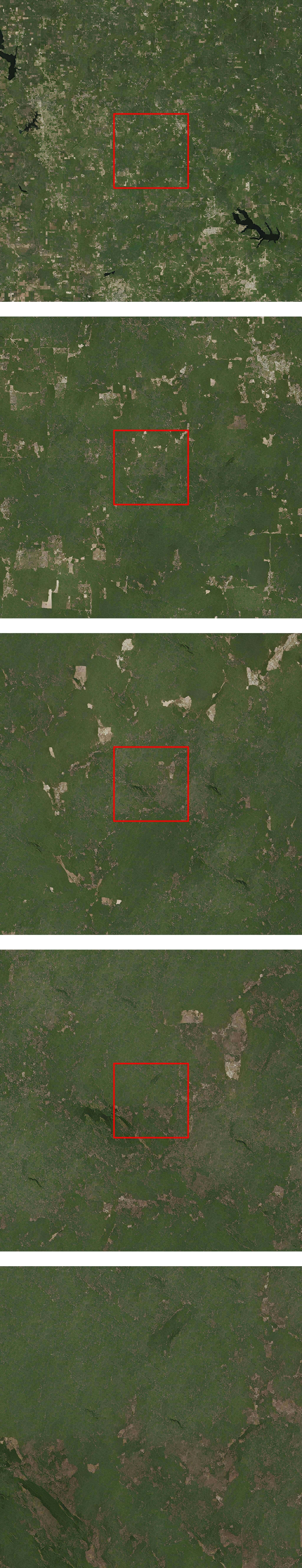}}
        }&
        
         \tikz{
        \node[draw=black, line width=.5mm, inner sep=0pt] 
        {\includegraphics[width=.14\linewidth]{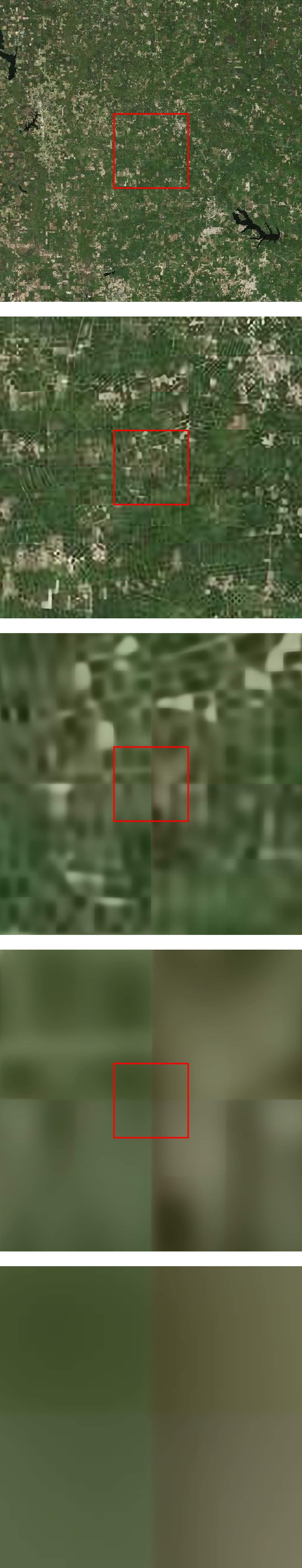}}
        }&
        \tikz{ 
        \node[draw=black, line width=.5mm, inner sep=0pt] 
        {\includegraphics[width=.14\linewidth]{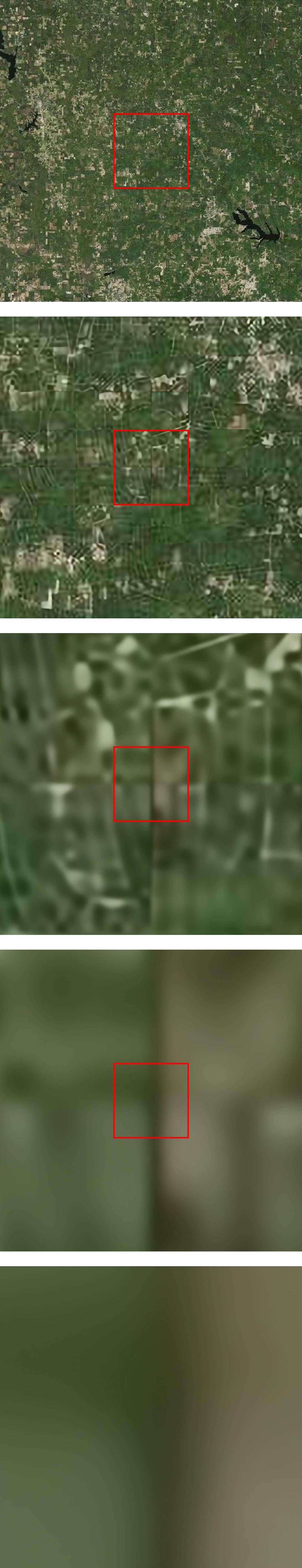}}
        }&
         \tikz{
        \node[draw=black, line width=.5mm, inner sep=0pt] 
        {\includegraphics[width=.14\linewidth]{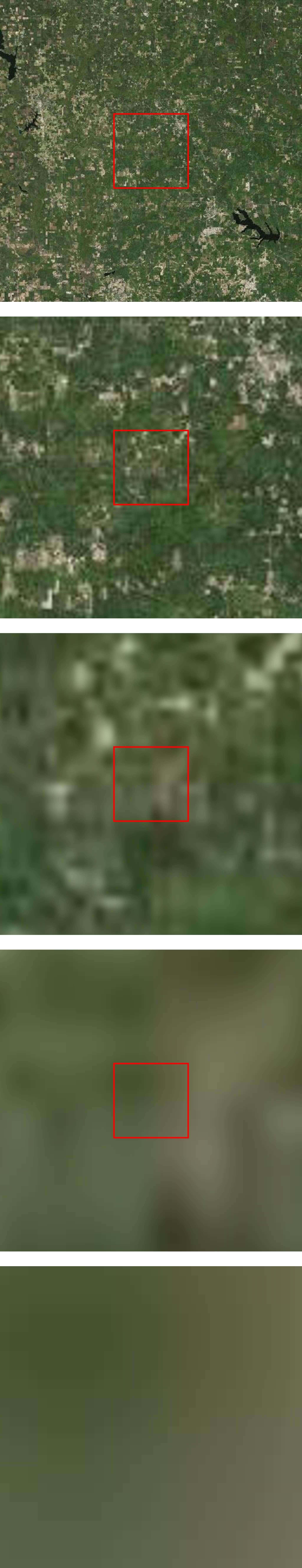}}
        }&

        \tikz{
        \node[draw=black, line width=.5mm, inner sep=0pt] 
        { \includegraphics[width=.14\linewidth]{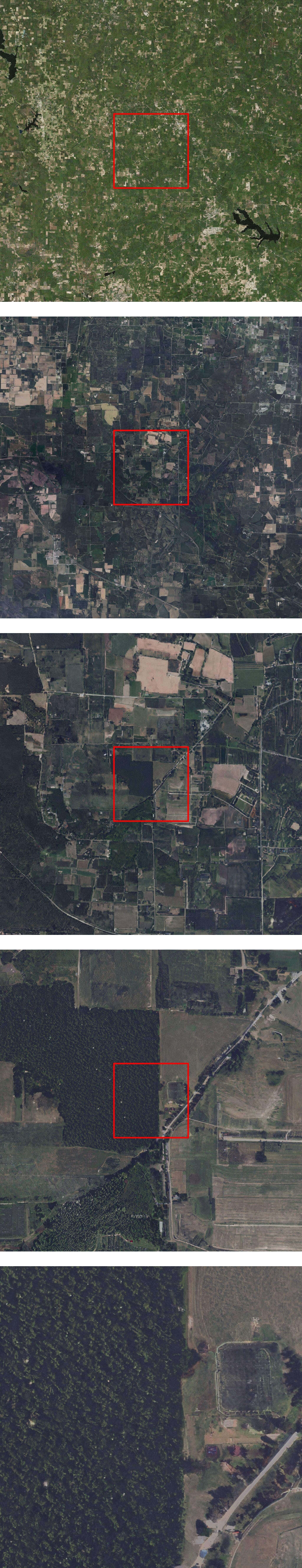}}
        }
    \end{tabular}
}
    \captionof{figure}{Qualitative Super-Resolution Comparison - Forest}
    \label{fig:qual_comp1}
\end{table*}

\begin{table*}[]
    \label{fig:baselines2} 
    \centering
    \resizebox{\linewidth}{!}{
\setlength{\tabcolsep}{0.2em} %
\renewcommand{\arraystretch}{1.}
    \begin{tabular}{cccccccc}
    
    GT &
     Stable Diff. &
     R-ESRGAN &
     HAT &
     LIIF &
    Interp. &
     Ours &
    \\
         \tikz{ 
        \node[draw=black, line width=.5mm, inner sep=0pt] 
        {\includegraphics[width=.14\linewidth]{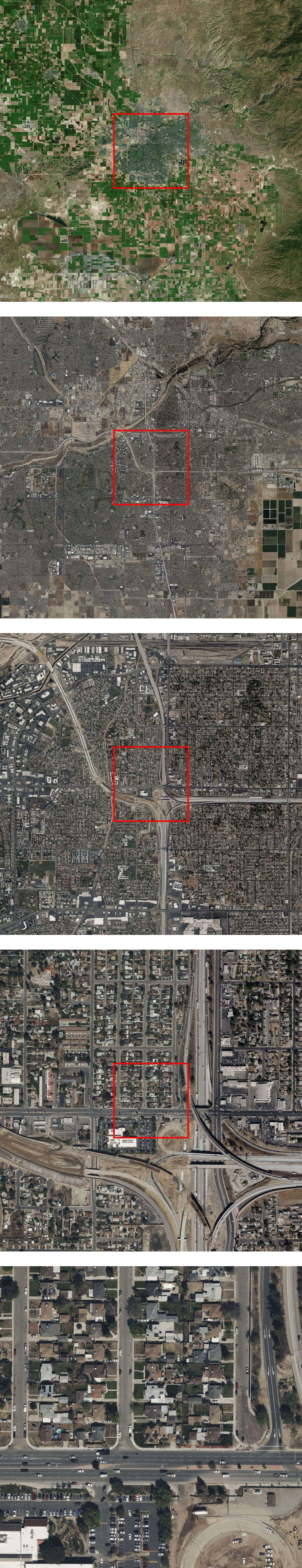}}
        }&
        \tikz{
        \node[draw=black, line width=.5mm, inner sep=0pt] 
        {\includegraphics[width=.14\linewidth]{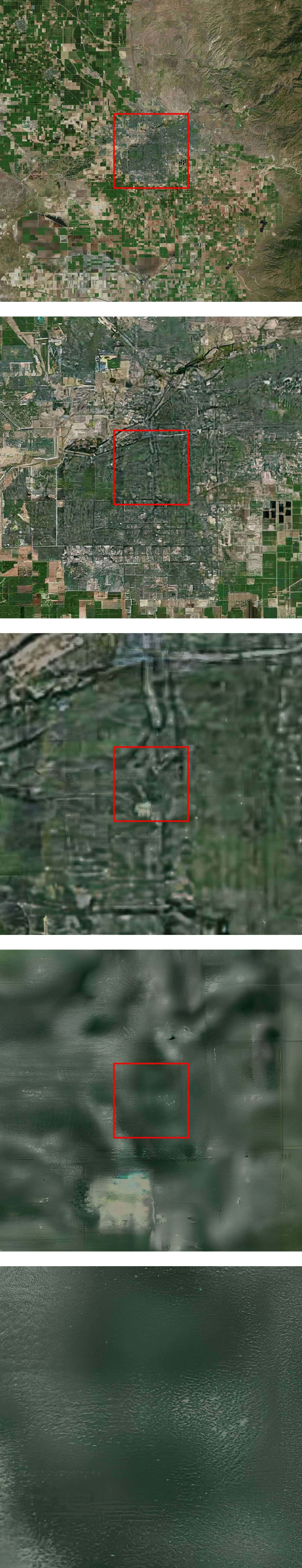}}
        }&
       
         \tikz{
        \node[draw=black, line width=.5mm, inner sep=0pt] 
        {\includegraphics[width=.14\linewidth]{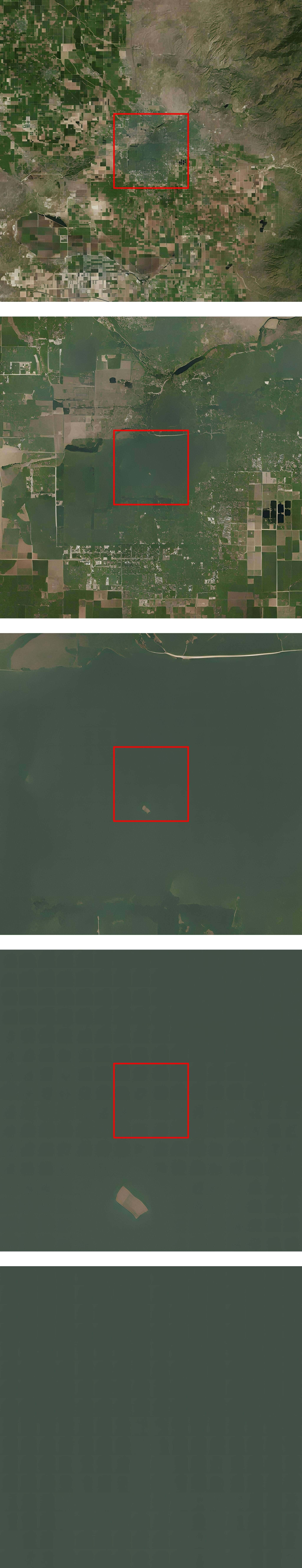}}
        }&
        
         \tikz{
        \node[draw=black, line width=.5mm, inner sep=0pt] 
        {\includegraphics[width=.14\linewidth]{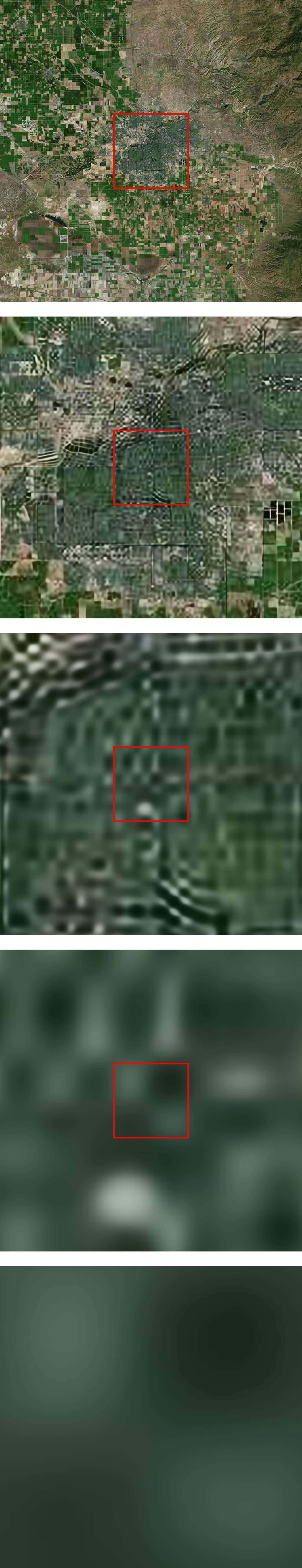}}
        }&
        \tikz{ 
        \node[draw=black, line width=.5mm, inner sep=0pt] 
        {\includegraphics[width=.14\linewidth]{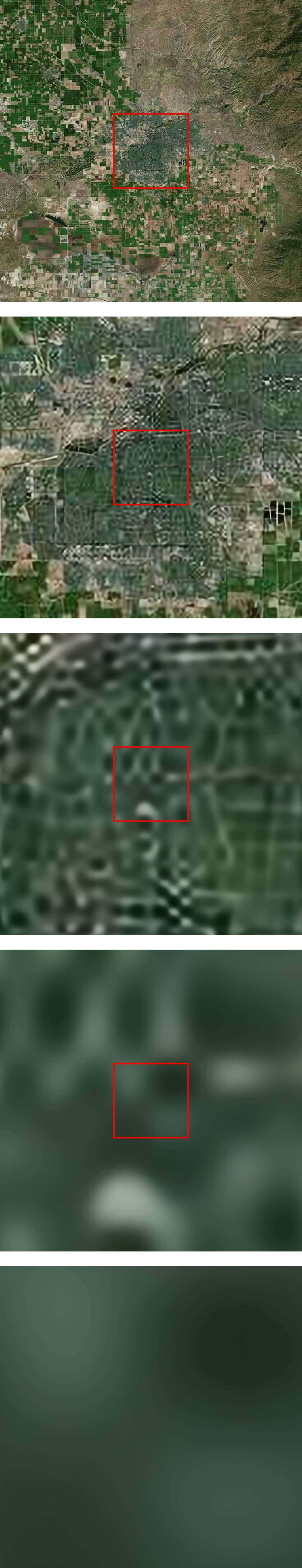}}
        }&
         \tikz{
        \node[draw=black, line width=.5mm, inner sep=0pt] 
        {\includegraphics[width=.14\linewidth]{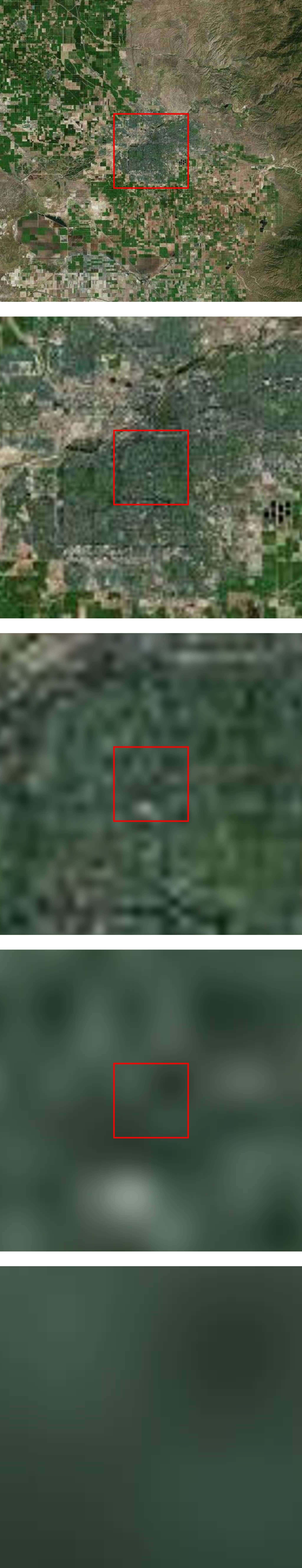}}
        }&

        \tikz{
        \node[draw=black, line width=.5mm, inner sep=0pt] 
        { \includegraphics[width=.14\linewidth]{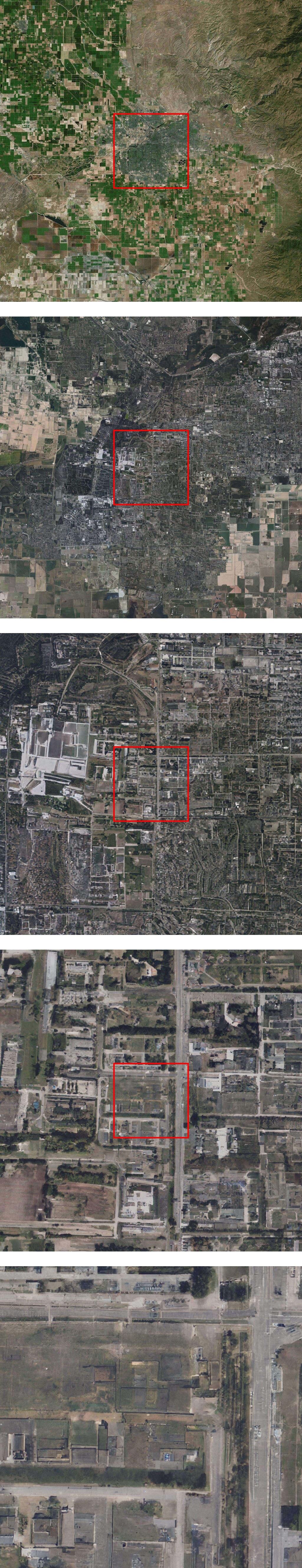}}
        }
    \end{tabular}
}
    \captionof{figure}{Qualitative Super-Resolution Comparison - Road}
    \label{fig:qual_comp2}
\end{table*}
\begin{table*}[]
    \label{fig:baselines3} 
    \centering
    \resizebox{\linewidth}{!}{
\setlength{\tabcolsep}{0.2em} %
\renewcommand{\arraystretch}{1.}
    \begin{tabular}{cccccccc}
    
    GT &
     Stable Diff. &
     R-ESRGAN &
     HAT &
     LIIF &
    Interp. &
     Ours &
    \\
         \tikz{ 
        \node[draw=black, line width=.5mm, inner sep=0pt] 
        {\includegraphics[width=.14\linewidth]{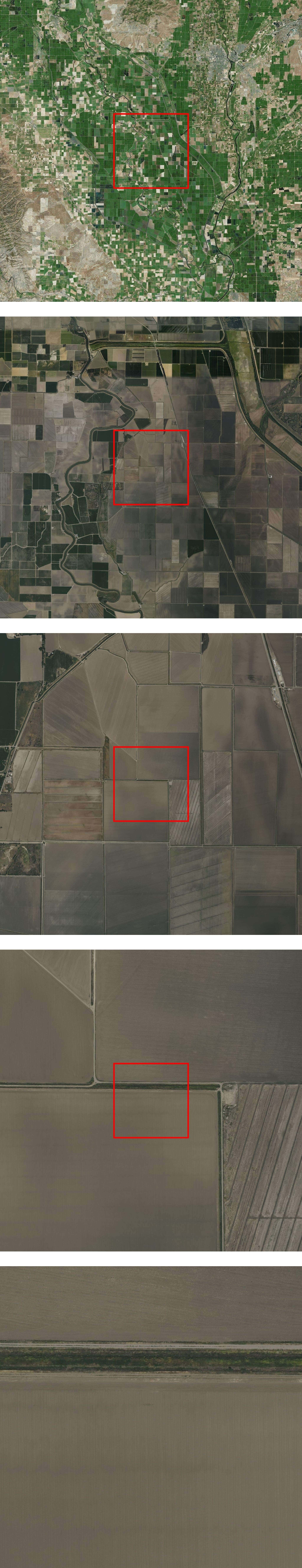}}
        }&
        \tikz{
        \node[draw=black, line width=.5mm, inner sep=0pt] 
        {\includegraphics[width=.14\linewidth]{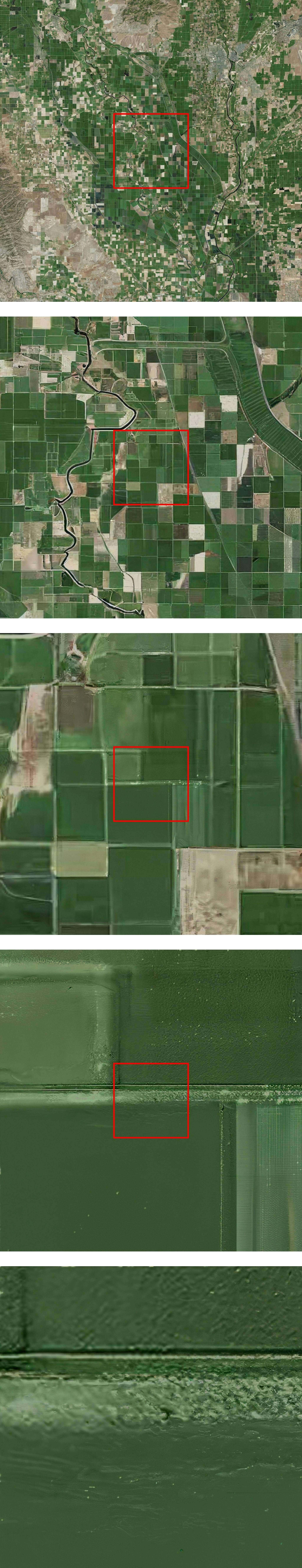}}
        }&
       
         \tikz{
        \node[draw=black, line width=.5mm, inner sep=0pt] 
        {\includegraphics[width=.14\linewidth]{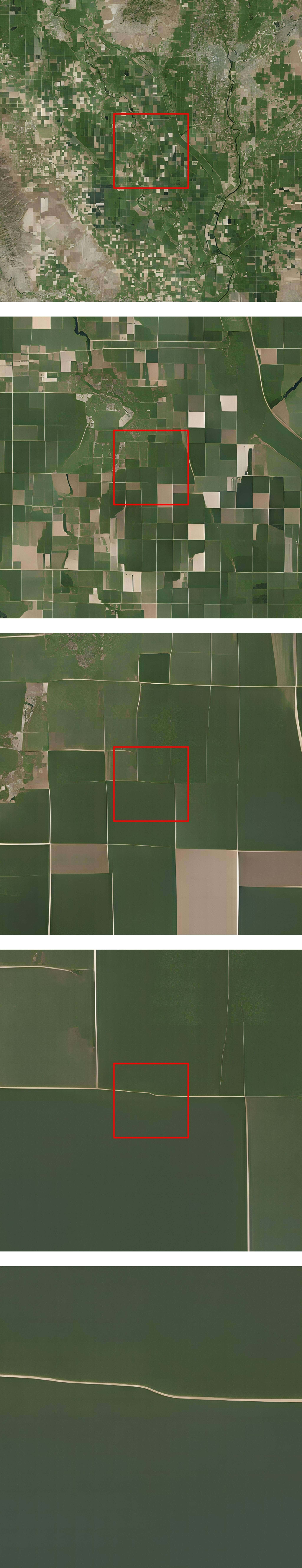}}
        }&
        
         \tikz{
        \node[draw=black, line width=.5mm, inner sep=0pt] 
        {\includegraphics[width=.14\linewidth]{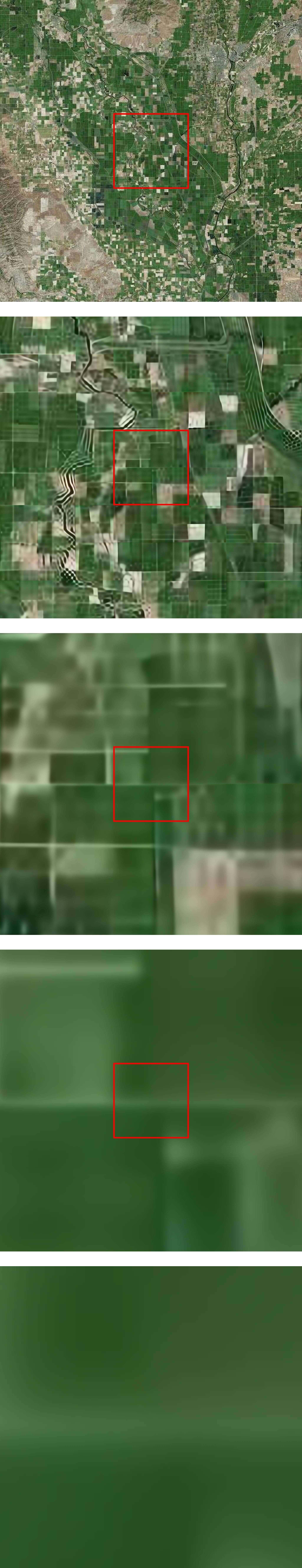}}
        }&
        \tikz{ 
        \node[draw=black, line width=.5mm, inner sep=0pt] 
        {\includegraphics[width=.14\linewidth]{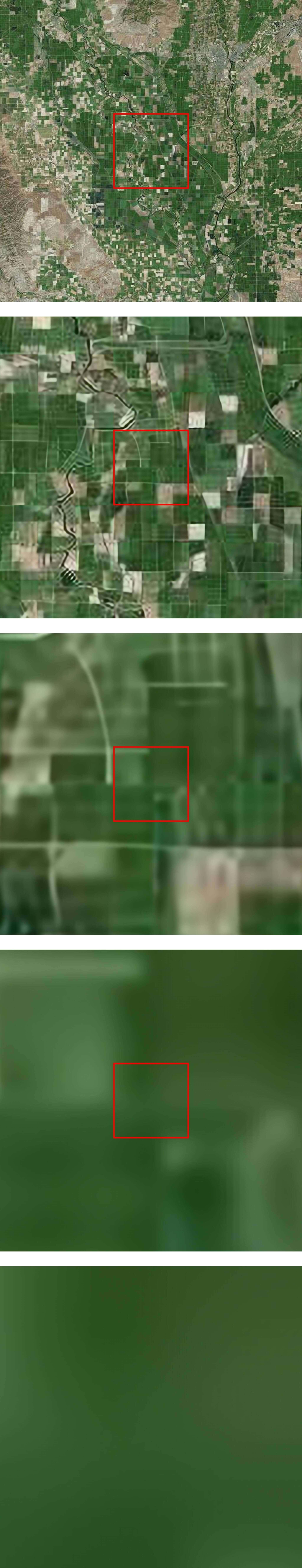}}
        }&
         \tikz{
        \node[draw=black, line width=.5mm, inner sep=0pt] 
        {\includegraphics[width=.14\linewidth]{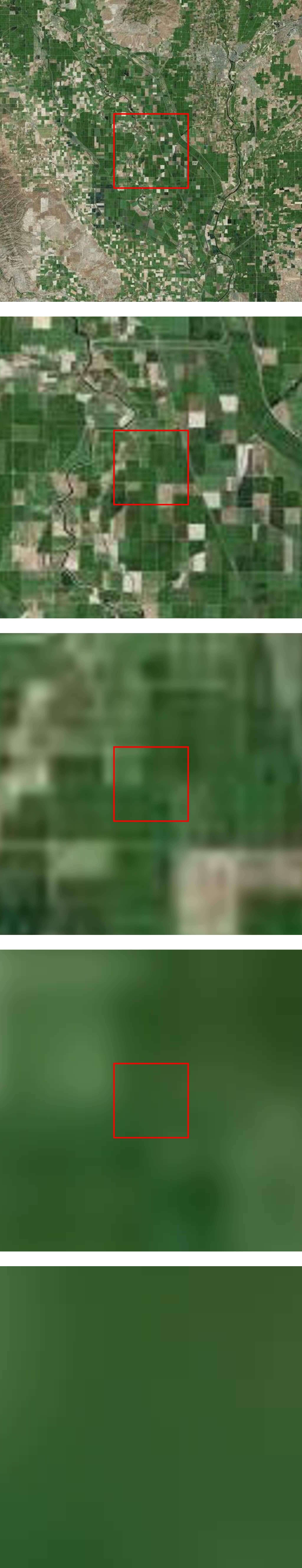}}
        }&

        \tikz{
        \node[draw=black, line width=.5mm, inner sep=0pt] 
        { \includegraphics[width=.14\linewidth]{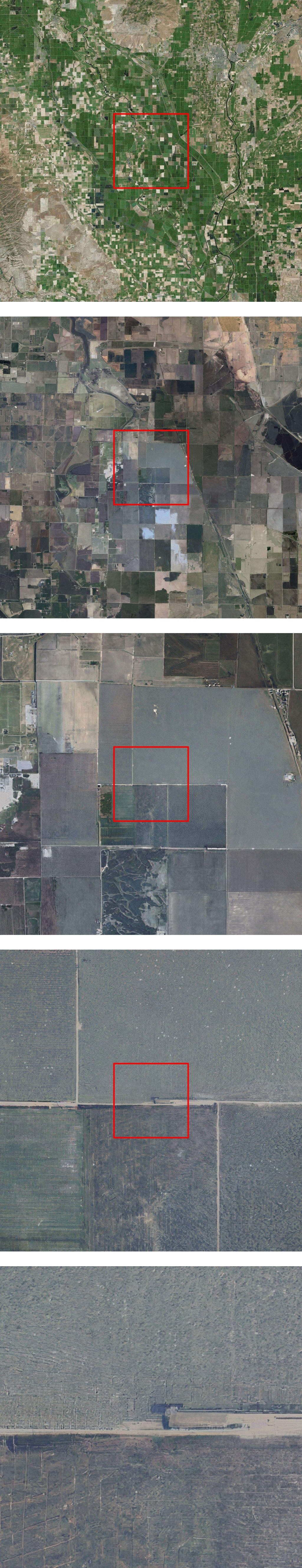}}
        }
    \end{tabular}
}
    \captionof{figure}{Qualitative Super-Resolution Comparison - Farm}
    \label{fig:qual_comp3}
\end{table*}

\subsection{Failure Cases}
\begin{figure*}[!htbp]
  \centering
  \includegraphics[angle=-90,width=0.9\textwidth]{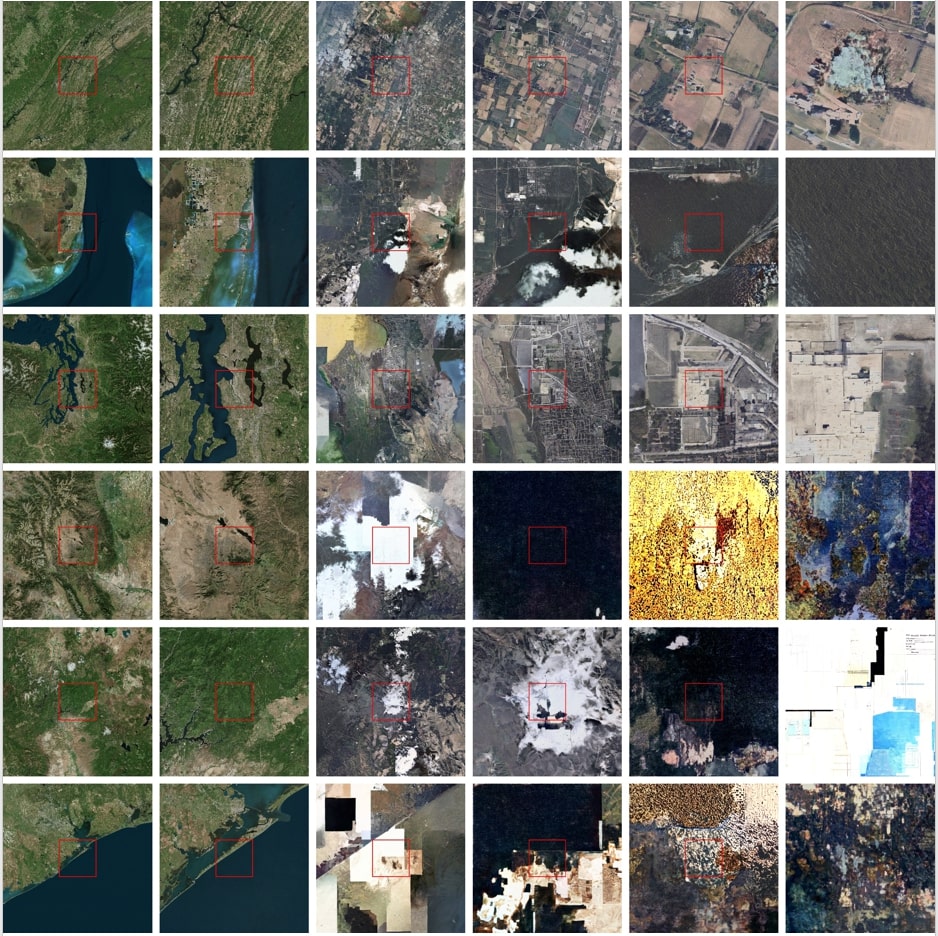}
  \caption{Example failure cases. Each column is a different location. Each row from top to bottom is our output from ground truth zoom 10 to generated zoom 12, and so on until generated zoom 20. Most failures occur due to divergence in the diffusion process caused by mostly monochrome tiles, which is somewhat alleviated in larger scale generations due to increased information propagation. Columns 1, 4, and 5 show failure cases over large bodies of water. Columns 2 and 3 represent failures due to monochrome tiles such as ice/water. Column 6 shows a failure at the final stage over a mostly textureless farm terrain for similar reasons.}
  \label{fig:fail}
\end{figure*}

The limitations of our model largely stem from the limitations of our dataset, as shown in the example failure cases in Fig. \ref{fig:fail}. Here, we see in the first, fourth, and fifth columns of the grid, the failure of our model to fully learn the real color shift of water in different kinds of bodies of water. However, we see that when small amounts of water are included (column 4), the remaining generations are not affected. The second and third columns of Fig. \ref{fig:fail} show the effects of zooming into clouds or snow, which occur mostly in mountainous terrain. The diffusion process also sometimes diverges due to strong text conditioning in the case of limited context. These tend to occur over mostly monochrome tiles such as water, shaded forests, ice/clouds, and other similar largely textureless patches, such as the farm from the last row of the last column in Fig. \ref{fig:fail}. Despite all of this, we observed that a large majority of these failures are alleviated by large-scale generations since more information can propagate through the Mixture of Diffusers overlapping tiled generation, as it gives more context to the model. Notably, one can observe significantly fewer failure cases in our interactive demo despite generating more tiles.

\clearpage

\end{document}